\documentclass[aerospace,article,accept,pdftex,moreauthors]{Definitions/mdpi} 

\firstpage{1} 
\pubvolume{1}
\issuenum{1}
\articlenumber{0}
\pubyear{2023}
\copyrightyear{2023}
\externaleditor{Academic Editor: Firstname Lastname}
\datereceived{12 September 2023} 
\daterevised{6 October 2023} 
\dateaccepted{10 October 2023} 
\datepublished{ } 
\hreflink{https://doi.org/} 

\Title{Learning to Predict 3D Rotational Dynamics from Images of a Rigid Body with Unknown Mass Distribution}

\TitleCitation{Learning to Predict 3D Rotational Dynamics from Images of a Rigid Body with Unknown Mass Distribution}

\Author{Justice J. Mason $^{1,2,*\dag}$, Christine Allen-Blanchette $^{1,*\dag}$, Nicholas Zolman $^{2}$, Elizabeth Davison $^{2}$ \linebreak and Naomi Ehrich Leonard $^{1}$}

\AuthorNames{Justice J. Mason, Christine Allen-Blanchette, Nicholas Zolman, Elizabeth Davison and Naomi Ehrich Leonard}

\AuthorCitation{Mason, J. J.; Allen-Blanchette, C.; Zolman, N.; Davison, E.; Leonard, N. E.}

\address{%
$^{1}$ \quad Department of Mechanical and Aerospace Engineering, Princeton University, Princeton, NJ 08544, USA; naomi@princeton.edu 
\\
$^{2}$ \quad The Aerospace Corporation, El Segundo, CA 90245, USA; nicholas.f.zolman@aero.org (N.Z.); elizabeth.davison@aero.org (E.D.)}

\corres{Correspondence: jjmason@princeton.edu (J.J.M.); ca15@princeton.edu (C.A.-B.)}

\firstnote{These authors contributed equally to this work.}

\abstract{In many real-world settings, image observations of freely rotating 3D rigid bodies may be available when low-dimensional measurements are not. However, the high-dimensionality of image data precludes the use of classical estimation techniques to learn the dynamics. The usefulness of standard deep learning methods is also limited, because an image of a rigid body reveals nothing about the distribution of mass inside the body, which, together with initial angular velocity, is what determines how the body will rotate. We present a physics-based neural network model to estimate and predict 3D rotational dynamics from image sequences. We achieve this using a multi-stage prediction pipeline that maps individual images to a latent representation homeomorphic to $\mathbf{SO}(3)$, computes angular velocities from latent pairs, and predicts future latent states using the Hamiltonian equations of motion. We demonstrate the efficacy of our approach on new rotating rigid-body datasets of sequences of synthetic images of rotating objects, including cubes, prisms and satellites, with unknown uniform and non-uniform mass distributions. Our model outperforms competing baselines on our datasets, producing better qualitative predictions and reducing the error observed for the state-of-the-art Hamiltonian Generative Network by a factor of 2.}


\newcommand{\jjm}{\color{black}}

\usepackage[utf8]{inputenc} 
\usepackage[T1]{fontenc}    
\usepackage{url}            
\usepackage{booktabs}       
\usepackage{amsfonts}       
\usepackage{nicefrac}       
\usepackage{microtype}      

\usepackage{epsfig}
\usepackage{graphicx}
\usepackage{subcaption}
\usepackage{amsmath}
\usepackage{amssymb}
\usepackage{listings}

\usepackage[ruled]{algorithm2e}
\usepackage{amsthm}
\usepackage{svg}
\usepackage{multirow}
\usepackage{siunitx}
\usepackage{graphics}
\usepackage{mwe}
\usepackage{caption}

 \usepackage[labelformat=simple]{subcaption}

\DeclareCaptionLabelFormat{subcaptionlabel}{\normalfont(\textbf{#2}\normalfont)}
\begin{document}

\section{Introduction}
\label{sec: intro}
{\jjm{The study and control of a range of systems can benefit from the means to }}predict the rotational dynamics of 3D rigid bodies that are only observed through images. {\jjm{A compelling example is the navigation and control of space robotic systems that interact with resident space objects (RSOs). RSOs are natural or designed freely rotating rigid bodies that orbit a planet or moon. Space robotic system missions that involve interaction with RSOs include collecting samples from an asteroid \citep{OSIRIS}, servicing a malfunctioning satellite \citep{OrbitService}, and removing active space debris \citep{DebrisRemoval}.  
A challenge is that space robotic systems may have limited information on the mass distribution of RSOs. However, they do typically have onboard cameras to observe sequences of RSO movement. Thus, learning to predict the dynamics of the RSOs from onboard images can make a difference for mission success.}} 

\begin{figure}[H]
   \includegraphics[width=0.99\textwidth]{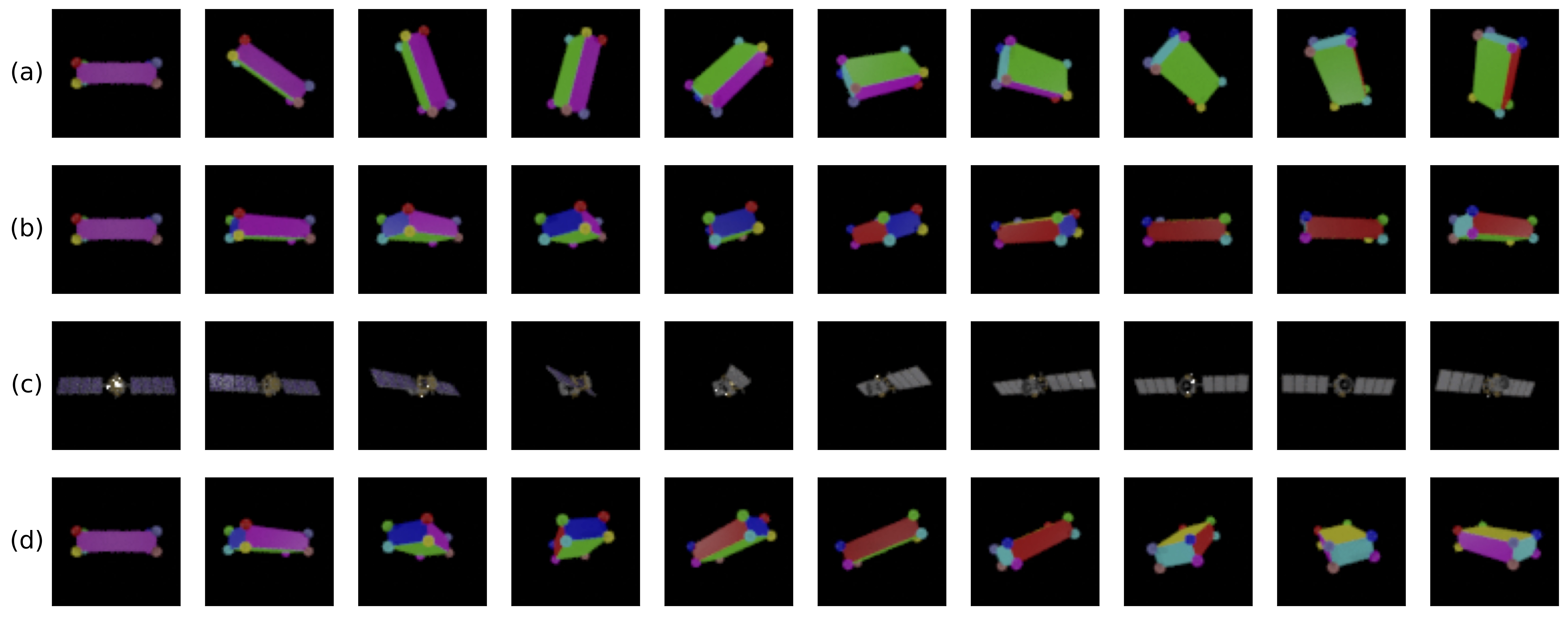}
  \caption{Simulations illustrating how mass distribution and initial angular velocity determine behavior. (\textbf{a}) Tumbling prism: uniform mass distribution ($\mathbf{J}$ 
 = $\mathbf{J}_1$) and initial angular velocity near an unstable solution. (\textbf{b}) 
Spinning prism: 
{$\mathbf{J} = \mathbf{J}_1$ and} initial angular velocity near a stable solution. 
(\textbf{c})
Spinning CALIPSO satellite: 
{$\mathbf{J} = \mathbf{J}_1$ 
and} same initial angular velocity as (\textbf{b}).  
(\textbf{d})
Wobbling prism: non-uniform mass distribution ($\mathbf{J}$ = $\mathbf{J}_3$) and same initial velocity as (\textbf{b}). 
}
  \label{fig: mass-distr-illustration}
\end{figure}

Whether a freely rotating 3D rigid body tumbles unstably or spins stably depends on the distribution of mass inside the body and the body's initial angular velocity (compare Figure~\ref{fig: mass-distr-illustration}a and Figure~\ref{fig: mass-distr-illustration}b). This means that to predict the body's rotational dynamics, it is not enough to know the external geometry of the body. That would be insufficient, for~instance, to~predict the different behavior of two bodies with the same external geometry and different internal mass distribution. Even if the bodies start at the same initial angular velocity, one body could tumble or wobble while the other spins stably (compare Figure~\ref{fig: mass-distr-illustration}b and Figure~\ref{fig: mass-distr-illustration}d).

{{Figure~\ref{fig: mass-distr-illustration} shows four simulations of a freely rotating rigid body that illustrate the role of mass distribution and initial velocity. The~distribution of mass determines $\mathbf{J} \in \mathbb{R}^{3 \times 3}$, where $\mathbf{J}$ is the \textit{moment-of-inertia matrix} for a 3D rigid body expressed with respect to the body-fixed frame, i.e.,~an orthonormal reference frame ${\cal B} = \{\mathbf{i}, \mathbf{j}, \mathbf{k}\}$ fixed to the body with origin at the body's center of mass (see Appendix~\ref{app:rigid_body} for details). Figure~\ref{fig: mass-distr-illustration}a--c all have the same moment-of-inertia matrix $\mathbf{J} = \mathbf{J}_1$, which corresponds to that of a rectangular prism with uniform mass distribution (see Table~\ref{table: mass-props} in Appendix~\ref{appendix: hyper-parameters}). Steady spin about the longest and shortest {\jjm{principal}} axes is stable and about the intermediate principal axis is unstable (see Appendix~\ref{app:rigid_body}). So, if the initial angular velocity is near the unstable solution, the body  tumbles (Figure~\ref{fig: mass-distr-illustration}a), whereas if it is near the stable axis, the body spins (Figure~\ref{fig: mass-distr-illustration}b). This is independent of the external geometry, which explains why the satellite in Figure~\ref{fig: mass-distr-illustration}c spins identically to the prism in Figure~\ref{fig: mass-distr-illustration}b. 
In Figure~\ref{fig: mass-distr-illustration}d, mass is non-uniformly distributed, such that $\mathbf{J} = \mathbf{J}_3$ (see Table~\ref{table: mass-props} in Appendix~\ref{appendix: hyper-parameters}) and the same initial velocity as in Figure~\ref{fig: mass-distr-illustration}b is no longer close to a stable solution, which explains why the prism wobbles.
}}

Predicting 3D rigid body rotational dynamics is possible if the body's mass distribution can be learned from observations of the body in motion. This is easier if the observations consist of low-dimensional data, e.g.,~measurements of the body's angular velocity and the rotation matrix that defines the body's orientation. It is much more challenging, however, if~the only available measurements consist of images of the body in motion, as~in the case of remote observations of a~satellite or asteroid or space debris.    

We address the challenge of learning and predicting 3D rotational dynamics from image sequences of a rigid body with unknown mass distribution and unknown initial angular velocity. To~do so we design a neural network model that leverages Hamiltonian structure associated with 3D rigid body dynamics. We show how our approach outperforms applicable methods from the existing~literature. 

Deep learning has proven to be an effective tool to learn dynamics from images. Previous work~\cite{ZhongLeonard2020, Toth2020HGN, Allen-Blanchette2020Lag} has made significant progress in using physics-based priors to learn dynamics from images of 2D rigid bodies, such as a pendulum. Learning dynamics of 3D rigid-body motion has also been explored with various types of input data~\cite{Byravan2017SE3, Valentin2020, duong21hamiltonian}. We believe our method is the first to use the Hamiltonian formalism to learn 3D rigid-body rotational dynamics from~images.

\begin{figure}[H]
{\captionsetup{position=bottom,justification=centering}
 \begin{subfigure}{0.48\textwidth}
     \includegraphics[width=\textwidth]{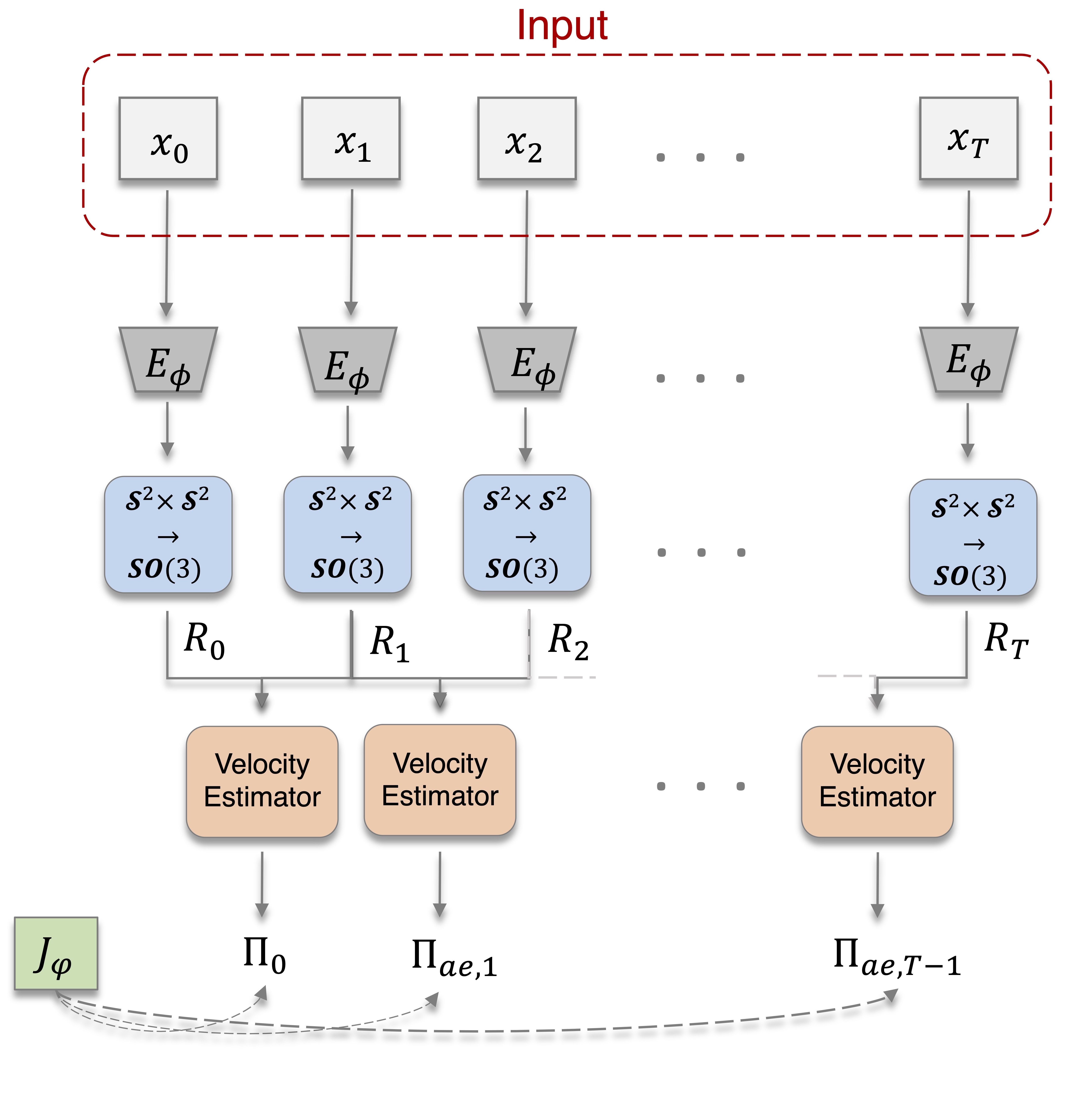}
     \caption{Encoding pipeline at~training}
     \label{fig:encode_train}
 \end{subfigure}
 \hfill
 \begin{subfigure}{0.4\textwidth}
     \includegraphics[width=\textwidth]{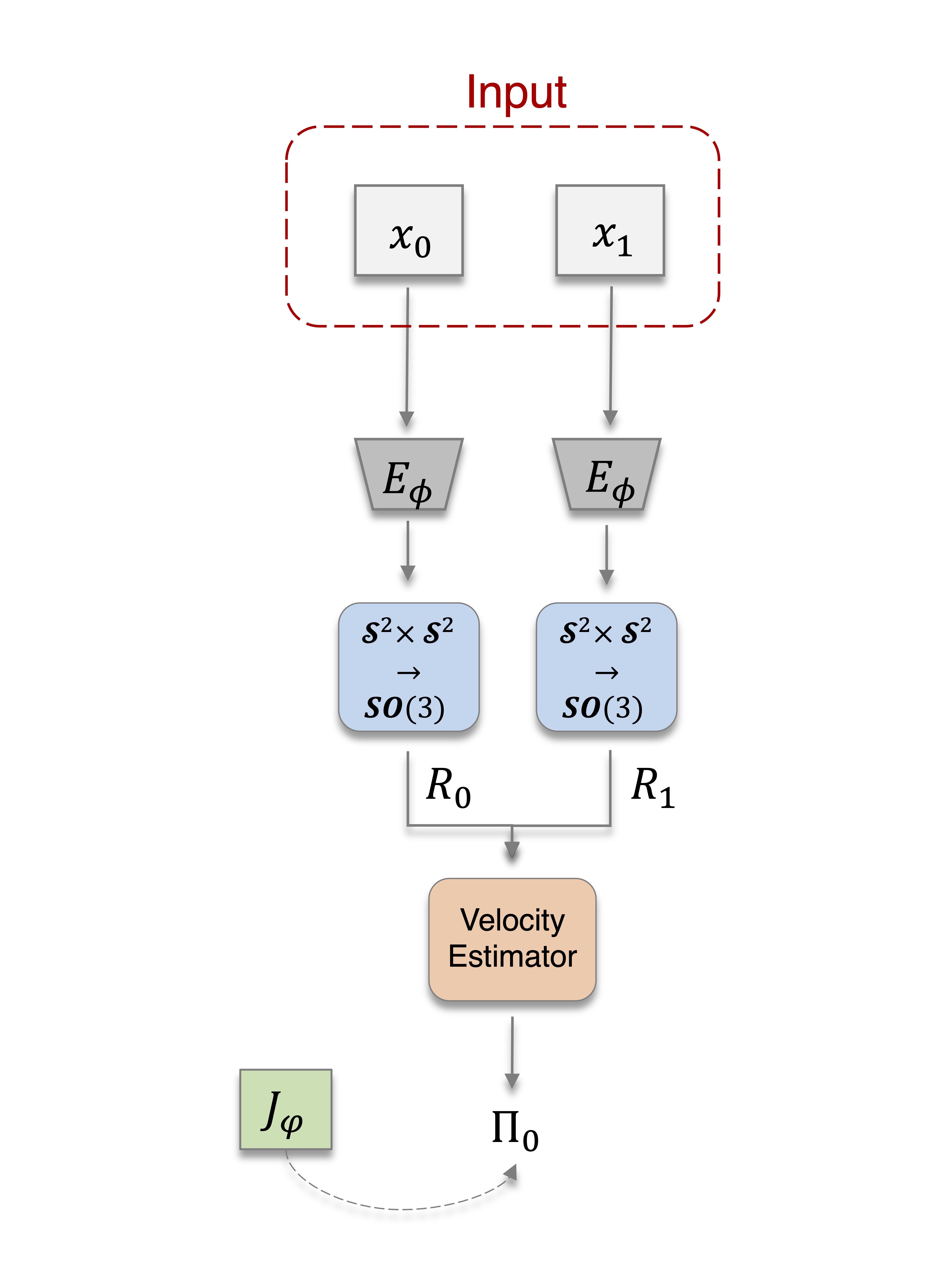}
     \caption{Encoding pipeline at~inference}
     \label{fig:encode_inference}
 \end{subfigure}
 
 \medskip
 \begin{subfigure}{0.48\textwidth}
     \includegraphics[width=\textwidth]{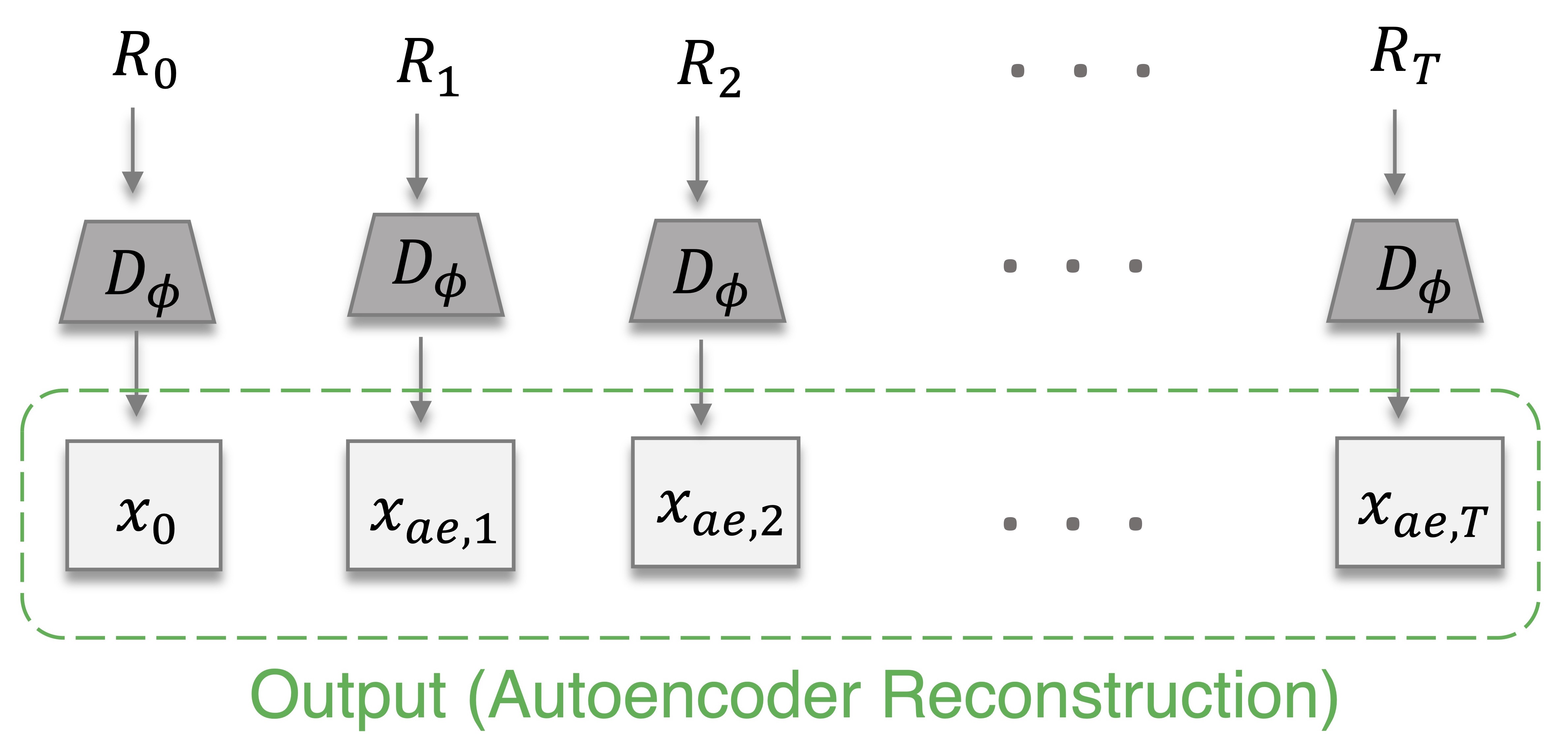}
     \caption{Auto-encoding~reconstruction}
     \label{fig:decode_recon}
 \end{subfigure}
 \hfill
 \begin{subfigure}{0.48\textwidth}
     \includegraphics[width=\textwidth]{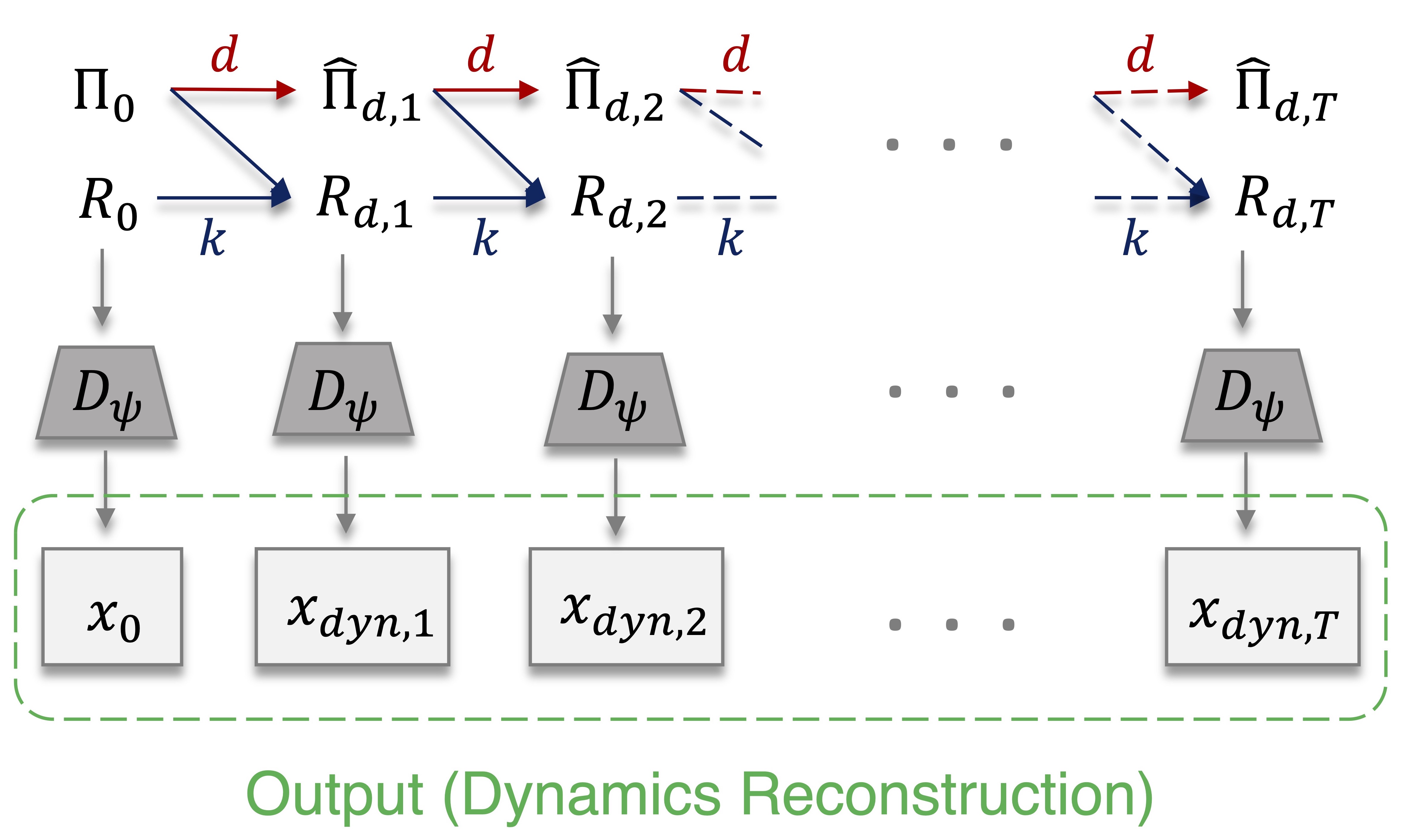}
     \caption{Dynamics-based~reconstruction}
     \label{fig:decode_pred}
 \end{subfigure}}
\vspace{5pt}

 \caption{A schematic of the model's forward pass at training time and inference. 
 {(\textbf{a}) Encoding pipeline at training; (\textbf{b}) encoding pipeline at inference; (\textbf{c}) decoding for auto-encoding reconstruction; and (\textbf{d}) dynamics prediction and decoding for dynamics-based reconstruction.} 
 }
 \label{fig: model_arch_train_inf}
\end{figure}

In this work, we introduce a model, with~architecture depicted in Figure~\ref{fig: model_arch_train_inf}, that (1) learns 3D rigid-body rotational dynamics from images, (2) predicts future image sequences in time, and~(3) generates a low-dimensional, interpretable representation of the latent state. {{During training, our model encodes a sequence of images (input) to a sequence of latent orientations (Figure~\ref{fig: model_arch_train_inf}a). The~sequence of orientations is processed by two pathways. In~one, the~sequence is decoded to a sequence of images which are used to compute the auto-encoding reconstruction loss (Figure~\ref{fig: model_arch_train_inf}c). In~the other, the~first element of the sequence is processed by the dynamics pipeline. The~resulting sequence is decoded to a sequence of images, which are used to compute the dynamics-based reconstruction loss (Figure~\ref{fig: model_arch_train_inf}d). During~inference, our model encodes a pair of images (input) to a single latent orientation (Figure~\ref{fig: model_arch_train_inf}b). This latent orientation is processed by the dynamics pipeline and decoding pipeline resulting in a predicted image sequence (Figure~\ref{fig: model_arch_train_inf}d).}}

Our model incorporates the Hamiltonian formulation of the dynamics as an inductive bias to facilitate learning the moment-of-inertia matrix, $\mathbf{J}_\varphi$, and~an auto-encoding map between images and {{the special orthogonal group $\mathbf{SO}(3) = \big\{\mathbf{R} \in \mathbb{R}^{3\times3} \vert\, \mathbf{R}^{T}\mathbf{R} = \mathbf{I}_{3},\, \det (\mathbf{R}) = + 1\big\}$. $\mathbf{SO}(3)$ represents the space of all 3D rotations:}} {{the orientation of the rigid body at time $t$ is described by the rotation matrix $\mathbf{R}(t) \in \mathbf{SO}(3)$ that maps points on the body from body frame coordinates to inertial frame coordinates at time $t$.}}

The efficacy of our approach is demonstrated through long-term image prediction on synthetic datasets. Due to the scarcity of appropriate datasets, we have created publicly available, synthetic datasets of rotating objects (e.g., cubes, prisms, and~satellites) applicable for evaluation of our model, as well as other tasks on 3D rigid-body rotation such as pose~estimation.

\section{Related~Work}
\label{sec: related work}
A growing body of work incorporates Hamiltonian and Lagrangian formalisms to improve the accuracy and interpretability of learned representations in neural network-based dynamical systems forecasting~\cite{Greydanus2019HNN, Chen2020SRNN, Cranmer2020LNN}. \citet{Greydanus2019HNN} predict symplectic gradients of a Hamiltonian system using a Hamiltonian parameterized by a neural network. They show that the Hamiltonian neural network {\jjm{(HNN)}} predicts the evolution of conservative systems better than a baseline black-box model. \citet{Chen2020SRNN} improve the long-term prediction performance of~\cite{Greydanus2019HNN} by minimizing the mean-squared error (MSE) between ground-truth and predicted state trajectories rather than one-step symplectic gradients. \citet{Cranmer2020LNN} propose parameterization of the system Lagrangian by a neural network arguing that momentum coordinates may be difficult to compute in some settings. Each of the aforementioned learn from sequences of phase-space measurements; our model learns from~images.

The authors of~\cite{ZhongLeonard2020, Allen-Blanchette2020Lag, Toth2020HGN} leverage Hamiltonian and Lagrangian neural networks to learn the dynamics of 2D rigid bodies (e.g., the~planar pendulum) from image sequences. \citet{ZhongLeonard2020} introduce a coordinate-aware {variational autoencoder} (VAE)\cite{Kingma2014} with a latent Lagrangian neural network (LNN) which learns the underlying dynamics and facilitates control. \citet{Allen-Blanchette2020Lag} use a latent LNN in an auto-encoding neural network to learn dynamics without control or prior knowledge of the configuration-space structure. \citet{Toth2020HGN} use a latent HNN in a VAE to learn dynamics without control, prior knowledge of the configuration-space structure or dimension. Similarly to \citet{Toth2020HGN}, we use a latent HNN to learn dynamics. Distinctly, however, we consider 3D rigid body dynamics and incorporate prior knowledge of the configuration-space structure to ensure interpretability of the learned~representations.

Others have considered the problem of learning 3D rigid-body dynamics~\cite{Byravan2017SE3, Valentin2020, duong21hamiltonian}. \citet{Byravan2017SE3} uses point-cloud data and action vectors (forces)
as inputs to a black-box neural network to predict the resulting $\mathbf{SE}(3)$
transformation matrix, which represents the motion of objects within the input scene. {{The special Euclidean group $\mathbf{SE}(3) = \big\{(\mathbf{R},\mathbf{r})  \vert\, \mathbf{R} \in \mathbf{SO}(3), \, \mathbf{r} \in \mathbb{R}^3\big\}$ represents the space of all 3D rotations and translations: the orientation and position of the rigid body at time $t$ is described by the rotation matrix and vector pair $(\mathbf{R}(t),\mathbf{r}(t)) \in \mathbf{SE}(3)$ that maps points on the body from body frame coordinates to inertial frame coordinates at time $t$.}} \citet{Valentin2020} create a novel symmetric matrix representation of $\mathbf{SO}(3)$ and incorporate it into a neural network to perform orientation prediction on synthetic point-cloud data and images. \citet{duong21hamiltonian} use low-dimensional measurement data (i.e., the~rotation matrix and angular momenta) to learn rigid body dynamics on $\mathbf{SO}(3)$ and $\mathbf{SE}(3)$ for~control. 

The combination of deep learning with physics-based priors allows models to learn dynamics from high-dimensional data such as images~\cite{ZhongLeonard2020, Toth2020HGN, Allen-Blanchette2020Lag}. However, as~far as we know, our method is the first to use the Hamiltonian formalism to learn 3D rigid-body rotational dynamics from~images.

\section{Background}
\label{sec: background}
\unskip
\subsection{The \texorpdfstring{$\mathcal{S}^2 \times \mathcal{S}^2$}{S2 x~S2} Parameterization of 3D Rotation Group \textbf{SO}(3)}\label{sec: param_so3}
The $\mathcal{S}^2 \times \mathcal{S}^2$ parameterization of the 3D rotation group $\mathbf{SO}(3)$ is a surjective and differentiable mapping with a continuous right inverse~\cite{HVAE2018}. Define the $n$-sphere:
    $\mathcal{S}^n = \big\{\mathbf{v}\in\mathbb{R}^{(n+1)} \vert \; v_1^2 + v_2^2 + \cdots + v_{n+1} ^2 = 1\big\}$.
The $\mathcal{S}^2 \times \mathcal{S}^2$ parameterization of $\mathbf{SO}(3)$ is given by
    $(\mathbf{u}, \mathbf{v}) \mapsto (\mathbf{w}_1, \mathbf{w}_2, \mathbf{w}_3)$
with
    $\mathbf{w}_1 = \mathbf{u}, \; \mathbf{w}_2 = \mathbf{v} - \mathbf{v}\langle \mathbf{u}, \mathbf{v}\rangle, \; \mathbf{w}_3 = \mathbf{w}_1 \times \mathbf{w}_2$, where $\mathbf{w}_i$ are renormalized to have unit~norm.

    Intuitively, this mapping constructs an orthonormal frame from the unit vectors $\mathbf{u}$ and $\mathbf{v}$ by Gram--Schmidt orthogonalization. The~right inverse of the parameterization is given by $(\mathbf{w}_1, \mathbf{w}_2, \mathbf{w}_3)\mapsto (\mathbf{w}_1, \mathbf{w}_2)$.
Other parameterizations of $\mathbf{SO}(3)$, such as the exponential map ($\mathfrak{so}(3)\mapsto \mathbf{SO}(3)$) and the quaternion map ($\mathcal{S}^3\mapsto \mathbf{SO}(3)$), do not have continuous inverses and therefore are more difficult to use in deep manifold regression~\cite{HVAE2018,levinson2020analysis,bregier2021deep,Zhou2019OnTC}.

\subsection{3D Rotating Rigid-Body~Kinematics}\label{subsec:so3-kine}
The orientation of a rotating 3D rigid body $\mathbf{R}(t) \in \mathbf{SO}(3)$ changing over time $t$ can be computed from \textit{body angular velocity 
} $\boldsymbol{\Omega}(t) \in \mathbb{R}^3$, i.e.,~the angular velocity of the body expressed with respect to the body frame ${\cal B}$, at~time $t \geq 0$ using the kinematic equations given by the time-rate-of-change of $\mathbf{R}(t)$ shown in Equation~\eqref{eq:Rdot}.
For computational purposes, 3D rigid-body rotational kinematics are commonly expressed in terms of the quaternion representation $\mathbf{q}(t) \in \mathcal{S}^3$  of the rigid-body orientation $\mathbf{R}(t)$. 
The kinematics~\eqref{eq:Rdot}, written in terms of quaternions~\cite{AndrleCrassidis}, are
\begin{equation}
    \label{eq: quat-kde}
    \frac{d\mathbf{q}(t)}{dt} = \textbf{Q}(\boldsymbol{\Omega}(t))\mathbf{q}(t), \;\;\;
    \textbf{Q}(\boldsymbol{\Omega}) = \begin{pmatrix}-\boldsymbol{\Omega}_{\times} & \boldsymbol{\Omega}\\ -\boldsymbol{\Omega}^{T} & 0\end{pmatrix},
\end{equation}
{{where $\boldsymbol{\Omega}_{\times}$ is the $3\times3$ skew-symmetric matrix defined by $(\boldsymbol{\Omega}_{\times}) \boldsymbol{y} = \boldsymbol{\Omega} \times \boldsymbol{y}$ for $\boldsymbol{y} \in \mathbb{R}^3$.}}

\subsection{3D Rigid-Body Dynamics in Hamiltonian~Form}
\label{subsec: LP-dyn}
The canonical Hamiltonian formulation derives the equations of motion for a mechanical system using only the symplectic form and a Hamiltonian function, which maps the state of the system to its total (kinetic plus potential) energy~\cite{GoldsteinCM}. This formulation has been used by several authors to learn unknown dynamics: the Hamiltonian structure (canonical symplectic form) is used as a physics prior and the unknown dynamics are uncovered by learning the Hamiltonian~\cite{Greydanus2019HNN, Zhong2020SymODEN, Toth2020HGN, Zhong2020Diss, Finci2020}. Consider a system with configuration space $\mathbb{R}^n$ and a choice of $n$ generalized coordinates that represent configuration. Let $\mathbf{z}(t)\in \mathbb{R}^{2n}$ represent the vector of $n$ generalized coordinates and their $n$ conjugate momenta at time $t$. Define the Hamiltonian function $\mathcal{H}:\mathbb{R}^{2n} \mapsto \mathbb{R}$ such that $\mathcal{H}(\mathbf{z})$ is the sum of the kinetic plus potential energy. Then, the equations of motion~\cite{GoldsteinCM, LeeLeokMcClamroch2018} derive as
\begin{equation}
    \label{eq:HamCan}
    \frac{d \mathbf{z}}{dt} = \Lambda_{\rm can} \nabla_{\mathbf{z}} \mathcal{H}(\mathbf{z}), \;\;\; \Lambda_{\rm can} = \begin{pmatrix} \boldsymbol{0}_n  & \mathbf{I}_n \\ -\mathbf{I}_n & \boldsymbol{0}_n \\ \end{pmatrix}
\end{equation}
where $\boldsymbol{0}_n \in \mathbb{R}^{n\times n}$ is the matrix of all zeros and $\Lambda_{\rm can}$ is the matrix representation of the \textit{canonical symplectic form}. 

The Hamiltonian equations of motion for a freely rotating 3D rigid body evolve on the six-dimensional space $T^*\mathbf{SO}(3)$, the~co-tangent bundle of $\mathbf{SO}(3)$. However, because~of rotational symmetry in the dynamics, i.e.,~the invariance of the dynamics of a freely rotating rigid body to the choice of inertial frame, the~Hamiltonian formulation of the dynamics can be reduced using the Lie--Poisson Reduction Theorem~\cite{MarsdenRatiu} to the space $\mathbb{R}^3 \sim \mathfrak{so}^*(3)$, the~Lie co-algebra of $\mathbf{SO}(3)$. These reduced Hamiltonian dynamics are equivalent to \eqref{eq:EulerEqn}, where the \textit{body angular momentum} is $\boldsymbol{\Pi}(t)\, { =\, \mathbf{J} \mathbf{\Omega}(t)} \in \mathfrak{so}^*(3)$ for $t\geq 0$. The~invariance can be seen by observing that the rotation matrix $\mathbf{R}(t)$, which describes the orientation of the body at time $t$, does not appear in \eqref{eq:EulerEqn}. $\mathbf{R}(t)$ is calculated from the solution of \eqref{eq:EulerEqn} using \eqref{eq:Rdot}. 

The reduced Hamiltonian $h:\mathfrak{so}^*(3) \mapsto \mathbb{R}$ for the freely rotating 3D rigid body (kinetic energy) is
\begin{equation}
    \label{eq: reduceHam}
    h(\boldsymbol{\Pi}) = \frac{1}{2} \boldsymbol{\Pi}\cdot \mathbf{J}^{-1}\, \boldsymbol{\Pi}.
\end{equation}

The reduced Hamiltonian formulation~\cite{MarsdenRatiu} is
\begin{equation}
    \label{eq:LiePoisson}
        \frac{d \boldsymbol{\Pi}}{dt} = \Lambda_{\mathfrak{so}^*(3)}(\boldsymbol{\Pi}) \nabla_{\boldsymbol{\Pi}}h(\boldsymbol{\Pi}), \;\;\; \Lambda_{\mathfrak{so}^*(3)}(\boldsymbol{\Pi}) = \boldsymbol{\Pi}_{\times},
\end{equation}
which can be seen to be equivalent to \eqref{eq:EulerEqn}. Equation \eqref{eq:LiePoisson}, called the \textit{Lie--Poisson equation}, generalizes the canonical Hamiltonian formulation. The~generalization allows for different symplectic forms, i.e.,~$\Lambda_{\mathfrak{so}^*(3)}$ instead of $\Lambda_{\rm can}$ in this case, each of which is only related to the latent space and symmetry. Our physics prior is the generalized symplectic form and learning the unknown dynamics means learning the reduced Hamiltonian. This is a generalization of the existing literature, where dynamics of canonical Hamiltonian systems are learned with the canonical symplectic form as the physics prior~\cite{Greydanus2019HNN, Cranmer2020LNN, Chen2020SRNN, Toth2020HGN}. Using the generalized Hamiltonian formulation allows extension of the approach to a much larger class of systems than those described by Hamilton's canonical equations, including rotating and translating 3D rigid bodies, rigid bodies in a gravitational field, multi-body systems, and~more.

\section{Materials and~Methods}
\label{sec: learn-so3}
In this section, we outline our approach for learning and predicting rigid-body dynamics from image sequences. The~multi-stage prediction pipeline maps individual images to an $\mathbf{SO}(3)$ latent space where angular velocities are computed from latent pairs. Future latent states are computed using the generalized Hamiltonian equations of motion \eqref{eq:LiePoisson} and a learned representation of the reduced Hamiltonian \eqref{eq: reduceHam}. Finally, the~predicted latent representations are mapped to images giving a predicted image~sequence. 

\subsection{Notation}
\label{subsec: notation}
$N$ denotes the number of image sequences in the dataset, and~$T+1$ is the length of each image sequence. Image sequences are written $\mathbf{x}_k = \{x_0^k, \dots, x_T^k\}$, 
sequences of latent rotation matrices are written $\mathbf{R}_k = \{R_0^k, \dots, R_T^k\}$ with $R_i^k\in\mathbf{SO}(3)$, and~quaternion latent sequences are written $\mathbf{q}_k = \{q_0^k, \dots, q_T^k\}$ with $q_i^k\in\mathcal{S}(3)$. Each element $y_i^k$ represents the quantity $y$ at time step $t = i$ for sequence $k$ from the dataset, where $k\in\{1, \dots, N\}$. Quantities generated with the learned dynamics are denoted with a hat (e.g., $\hat{q}$).

\subsection{Embedding Images to an \textbf{SO}(3) Latent Space}
\label{subsec: embedding}

{{In the first stage of our prediction pipeline, we embed image observations of a freely rotating rigid body to a low-dimensional latent representation to facilitate computation of the dynamics. The~latent representation is constrained to have the same $\mathbf{SO}(3)$ structure as the configuration space of the rigid body, making learned representations interpretable and~compatible with the equations of motion. Our embedding network $\Phi$ is given by the composition of functions $\Phi:=f\circ\pi\,\circ E_\phi: \mathcal{I}\mapsto\mathbf{SO}(3)$. The~convolutional encoding neural network $E_\phi:\mathcal{I}\mapsto \mathbb{R}^6$ parameterized by $\phi$ maps image observations from image space $\mathcal{I}$ to a vector $\mathbf{z}\in\mathbb{R}^6$. The~projection operator $\pi:\mathbb{R}^6\mapsto\mathcal{S}^2\times\mathcal{S}^2$ decomposes the vector $z$ into the vectors $\mathbf{u},\mathbf{v}\in\mathbb{R}^3$ and normalizes them, i.e.,~$\pi(\mathbf{z}) = (\mathbf{u}/\|\mathbf{u}\|,\, \mathbf{v}/\|\mathbf
{v}\|)$. Finally, the~function $f:\mathcal{S}^2\times \mathcal{S}^2\mapsto\mathbf{SO}(3)$ maps the normalized vectors $\mathbf{u}$ and $\mathbf{v}$ to the configuration space using the surjective and differentiable $\mathcal{S}^2\times \mathcal{S}^2$ parameterization of $\mathbf{SO}(3)$ (see Section~\ref{sec: param_so3}).}}

\subsection{Computing Dynamics in the Latent~Space}
\label{subsec:learninglatentdyn}
In the second stage of our prediction pipeline, we compute the dynamics of the freely rotating rigid body using a Hamiltonian with a learned moment-of-inertia tensor, $\mathbf{J}_{\varphi}$.  The~moment-of-inertia tensor, $\mathbf{J}_{\varphi}$, is parameterized by the vectors $\boldsymbol{\varphi}_1, \boldsymbol{\varphi}_2 \in \mathbb{R}^3$, representing the diagonal and off-diagonal components of the matrix, and~computed using the Cholesky decomposition~\cite{Lin_2019}. 

To compute the dynamics, we first construct an initial condition $(R_0^k, \Pi_0^k)\in T^{*}\mathbf{SO}(3)$. Given the sequential pair $(R_0^k, R_1^k)=(\Phi(x_0^k), \Phi(x_1^k))$, we perform this in two steps. First, we compute the angular velocity $\Omega_0^k$ by Algorithm~\ref{alg: BAVE}. Then, we compute the angular momentum by the matrix product of the learned moment-of-inertia and angular velocity, i.e.,~$\Pi_0^k = \mathbf{J}_\varphi\Omega_0^k$. With~the initial condition $(R_0^k, \Pi_0^k)$, subsequent angular momenta $\{\hat{\Pi}_i^k\}^T_{i=1}$ are computed using the Lie--Poisson Equation~\eqref{eq:LiePoisson} and the reduced Hamiltonian formed using the learned momentum-of-inertia $\mathbf{J}_{\varphi}$. We integrate the Lie--Poisson equations forward in time using a Runge--Kutta fourth-order (RK45) numerical~solver~\cite{RungeKutta}. 
\newpage

\begin{algorithm*}[H]
\caption{An algorithm to calculate the body angular velocity given two sequential orientation matrices and the time step in between~them.}\label{alg: BAVE}
\KwData{$R_{t}, R_{t+1}, \Delta t$}
\KwResult{$\Omega_{t} = R_{t}\big(\frac{\theta}{\Delta t}\mathbf{u}\big)$}
$R_\text{prod} \gets R_{t+1}\big(R_{t}^T\big)$\;
$\mathbf{u}_{\times} \gets R_\text{prod}^T - R_\text{prod}$\;
$\theta \gets \arccos{\big(\frac{\text{Trace}(R_\text{prod}) - 1}{2}\big)}$\;
\eIf{$\mathbf{u}_{\times} = \mathbf{0}$}{
  \eIf{$\theta = 0$}{
    $\mathbf{u} \gets \begin{pmatrix} 1, 1, 1\end{pmatrix}$\;
    $\mathbf{u} \gets \textit{normalize}(\mathbf{u})$\;
  }{\If{$\theta = \pi$}{
      $\mathbf{u} \gets \textit{column}\big(R_\text{prod} + \mathbb{I}_3\big)$\;
    $\mathbf{u} \gets \textit{normalize}(\mathbf{u})$\;
    }
  }
}{\If{$\mathbf{u}_{\times} \neq \mathbf{0}$}{
    $\mathbf{u} \gets \textit{skew}^{-1}(\mathbf{u})$\;
    $\mathbf{u} \gets \textit{normalize}(\mathbf{u})$\;
    }
} 
\end{algorithm*}

\vspace{12pt}

Subsequent rotations $\{\hat{R}_i^k\}_{i=1}^T$ are computed in two steps. First, we compute the sequence of quaternions $\{\hat{q}_i^k\}^T_{i=1}$ by Equation~\eqref{eq: quat-kde}, using the quaternion representation $q_0^k$ of the initial rotation $R_0^k$ and the initial angular velocity $\Omega_0^k$. We integrate Equation~\eqref{eq: quat-kde} forward in time using an RK45 solver with a normalization step~\cite{AndrleCrassidis} that ensures elements of the resulting sequence are valid quaternions. Then, we transform the sequence of quaternions $\{\hat{q}_i^k\}^T_{i=1}$ to a sequence of rotations rotations $\{\hat{R}_i^k\}_{i=1}^T$ using a modified Shepperd's algorithm~\cite{Markley}.

\subsection{Decoding \textbf{SO}(3) Latent States to Images}
\label{subsec: decoding_pipeline}
In the final stage of our prediction pipeline, we decode the sequence of $\mathbf{SO}(3)$ latent states produced by the dynamics pipeline to a sequence of images (see Figure~\ref{fig: model_arch_train_inf}d). Our decoding network $\Psi$ is given as the composition of functions $\Psi:=D_\psi\circ\pi^{-1}\circ f^{-1}:\mathbf{SO}(3) \mapsto\mathcal{I}$, where the convolutional decoding network $D_\psi:\mathbb{R}^6\mapsto \mathcal{I}$ parameterized by $\psi$ maps a vector $\mathbf{z}=(\mathbf{u},\mathbf{v})$, where $(\mathbf{u},\mathbf{v})\in\mathcal{S}^2 \times \mathcal{S}^2$, to~the image space $\mathcal{I}$.

\subsection{Training~Methodology}
\label{subsec: train_method} 
In this section, we describe the loss functions used to optimize our model: 
 the auto-encoding reconstruction loss ($\mathcal{L}_\text{ae}$), the~dynamics-based reconstruction loss ($\mathcal{L}_\text{dyn}$), {{the latent orientation loss ($\mathcal{L}_\text{latent, R}$), and~latent momentum loss ($\mathcal{L}_\text{latent, $\Pi$}$).}} 
$\mathcal{L}_\text{ae}$ ensures the embedding to $\mathbf{SO}(3)$ is sufficiently expressive to represent the entire image dataset, and~$\mathcal{L}_\text{dyn}$ ensures correspondence between the input image sequences and the images sequences produced by the learned dynamics. The~latent loss functions, $\mathcal{L}_\text{latent, R}$ and $\mathcal{L}_\text{latent, $\Pi$}$, ensure consistency between the latent states produced by the encoding pipeline and those produced by the dynamics~pipeline. 

For notational convenience, we denote the encoding pipeline $\mathcal{E}: \mathcal{I}\mapsto\mathcal{S}^3$ and the decoding pipeline $\mathcal{D}: \mathcal{S}^3\mapsto\mathcal{I}$. {{Quantities computed in the encoding pipeline use subscript ae (e.g., ${R^k_\text{ae}}_i$), while those computed in the dynamics pipeline use subscript dyn (e.g., ${R^k_\text{dyn}}_i$).}}

\subsubsection{Reconstruction Losses}
\label{subsec:recon_losses}
 The auto-encoding reconstruction loss is the mean squared error (MSE) between the ground-truth image sequence and the reconstructed image sequence without dynamics:
\begin{equation*}
\label{eq: loss-ae}
    \mathcal{L}_\text{ae} =  \frac{1}{NT}\sum_{k=1}^{N}\sum_{i=0}^{T-1} \big\lVert\, { x_i^k} - (\mathcal{D}\circ\mathcal{E})\big({ x_i^k}\big)\,\big\rVert_{2}^{2}\,.
\end{equation*}

The dynamics-based reconstruction loss is the MSE between the ground-truth image sequence and the image sequence produced by the dynamics pipeline:
\begin{equation*}
\label{eq: dynamics-loss}
    \mathcal{L}_\text{dyn} = \frac{1}{NT}\sum_{k=1}^{N}\sum_{i=1}^{T} \big\lVert\, { x_i^k} - \mathcal{D}\, \big({q^k_\text{dyn}}_i\big)\,\big\rVert_{2}^{2}\, .
\end{equation*}

\subsubsection{Latent Losses}
\label{subsec:latent_losses}
{{We define ${\mathcal{L}_\text{latent}}_R$ as}} the $\mathbf{SO}(3)$ distance~\cite{GoldsteinCM} between the $3\times3$ identity matrix and right-difference of orientations produced in the encoding pipeline and the orientations produced in the dynamics pipeline:
\begin{equation*}
\label{eq: latent-R-loss}
    {\mathcal{L}_\text{latent}}_R =
    \frac{1}{NT}\sum_{k=1}^{N}\sum_{i=1}^{T}
    \big\lVert\, \mathbf{I}_{3}\, -\,
    \big({R^k_\text{ae}}_i\big)^T\, {R^k_\text{dyn}}_i
    \big\rVert_F^2\, .
\end{equation*}

We define { ${\mathcal{L}_\text{latent}}_\Pi$ as} the MSE between the angular momenta estimated in the encoding pipeline and the angular momenta computed in the dynamics pipeline (see Figure~\ref{fig: model_arch_train_inf}):
\begin{equation*}
    \label{eq: latent-pi-loss}
{\mathcal{L}_\text{latent}}_\Pi\, =\, \frac{1}{NT}\sum_{k=1}^{N}\sum_{i=1}^{T} \big\lVert {\Pi^k_\text{ae}}_i\, -\, {\Pi^k_\text{dyn}}_i\big\rVert_{2}^{2}\, .
\end{equation*}

{{The hyperparameters we use to train our model are given in Table~\ref{table: train_params} in Appendix~\ref{appendix: training_params}. We train our model for 500 epochs, on~a single NVIDIA A100 SXM4 GPU. Our training time is approximately 12 hours, and~our inference time is approximately 300 milliseconds.}}
\subsection{3D Rotating Rigid-Body~Datasets} 
\label{subsec: data}
To evaluate our model, we introduce six synthetic datasets of freely rotating objects. Previous efforts in learning dynamics from images~\cite{Greydanus2019HNN, Toth2020HGN, Allen-Blanchette2020Lag, ZhongLeonard2020} consider only 2D planar systems (e.g., the~simple pendulum, Acrobot, and~cart-pole); existing datasets of freely rotating rigid bodies in 3D such as SPEED~\cite{SPEED1, SPEED2}, SPEED+ \cite{park2022speedplus}, and~URSO~\cite{URSO}, contain random image-pose pairs rather than sequential pairs needed for video prediction and dynamics extraction. Our datasets showcase the rich dynamical behaviors of 3D rotational dynamics through images, and~can be used for 3D dynamics learning tasks. {{Specifically, we introduce the following five datasets (see Table~\ref{table: mass-props} in Appendix~\ref{appendix: hyper-parameters} for moment-of-inertia matrices)}}:

\begin{itemize}
    \item \textbf{Uniform mass density cube}: a multi-colored cube of uniform mass density;
    \item \textbf{Uniform mass density prism}: a multi-colored rectangular prism with uniform mass density;
    \item \textbf{Non-uniform mass density cube}: a multi-colored cube with non-uniform mass density;
    \item \textbf{Non-uniform mass density prism}: \textls[-15]{a multi-colored prism with non-uniform mass density; }
    \item \textbf{Uniform density synthetic-satellites}: renderings of CALIPSO and CloudSat satellites with uniform mass density.
\end{itemize}

For each dataset, $N=1000$ trajectories are created. Each trajectory consists of an initial condition $\mathbf{x}_0 = (\mathbf{R}_0, \mathbf{\Pi}_0)$ that is integrated forward in time using a Python-based Runge--Kutta solver for $T=100$ time steps with spacing $\Delta t = 10^{-3}$. Initial conditions are chosen such that
$(\mathbf{R}_0, \mathbf{\Pi}_0) \sim \text{Uniform}\left(\mathbf{SO}(3) \times  S^2\right)$ with $\mathbf{\Pi}_0$ scaled to have $\lVert \mathbf{\Pi}_0 \rVert_2 = 50$. The~orientations $\hat{\mathbf{q}}$ from the integrated trajectories are passed to Blender~\cite{Blender} to render images of $28 \times 28$ pixels (as shown in Figure~\ref{fig: mass-distr-illustration}).

{{The synthetic image datasets are generated using Blender~\cite{Blender} with ideal and fixed lighting conditions. Models trained on this dataset may exhibit sensitivity to variations in lighting conditions, and~may not generalize to real~data.}}

\section{Results}
\label{sec: results}

\begin{figure}[H]
  \includegraphics[width=0.995\textwidth]{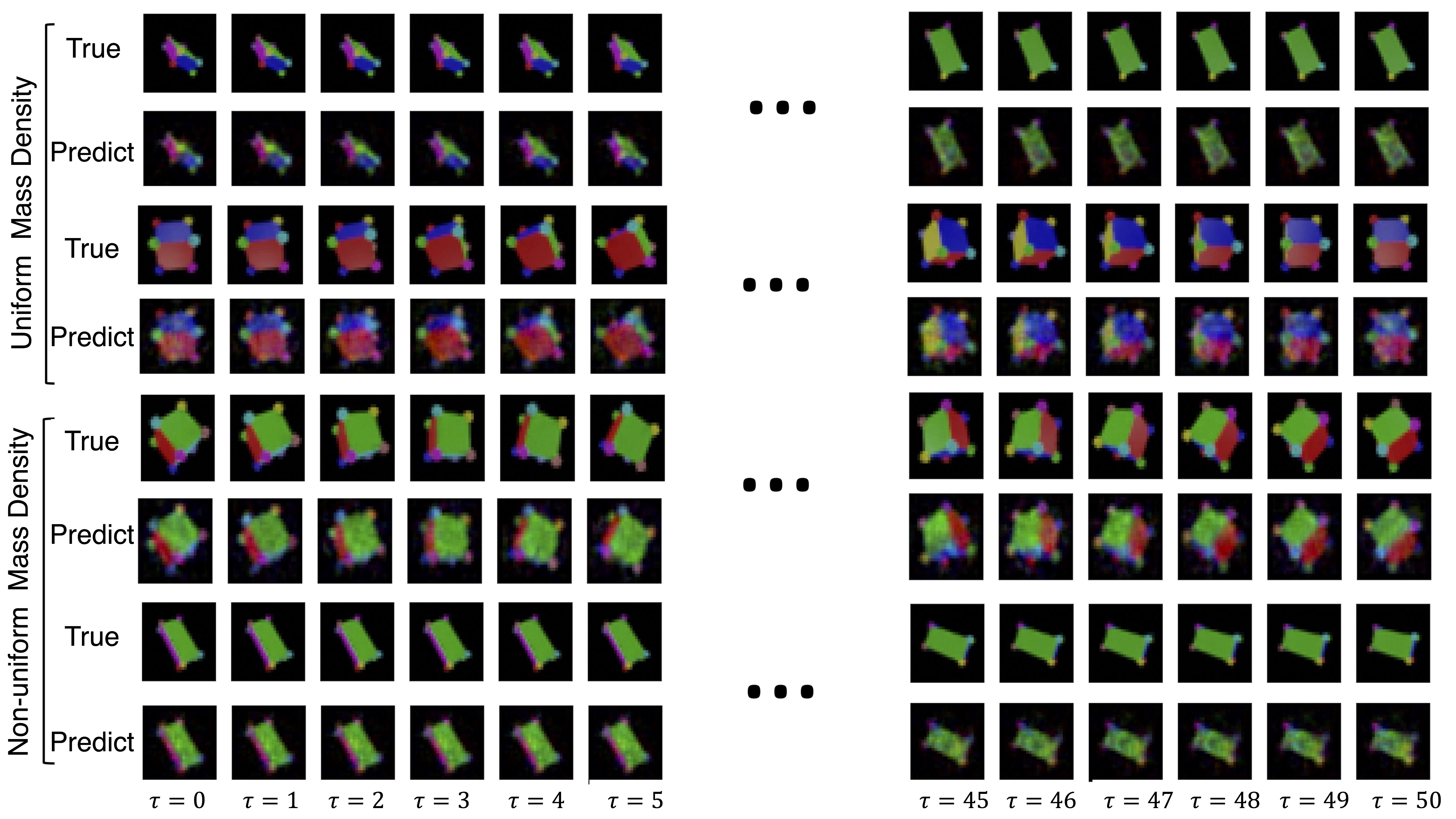}
  \caption{Predicted sequences for uniform and non-uniform mass density cube and prism datasets given by our model. The~figure shows predicted images at time steps $\tau = $ 0 to 5 and $\tau =$ 45 to 50.}
  \label{fig: prediction}
\end{figure}

\begin{figure}[H]
  \includegraphics[width=0.995\textwidth]{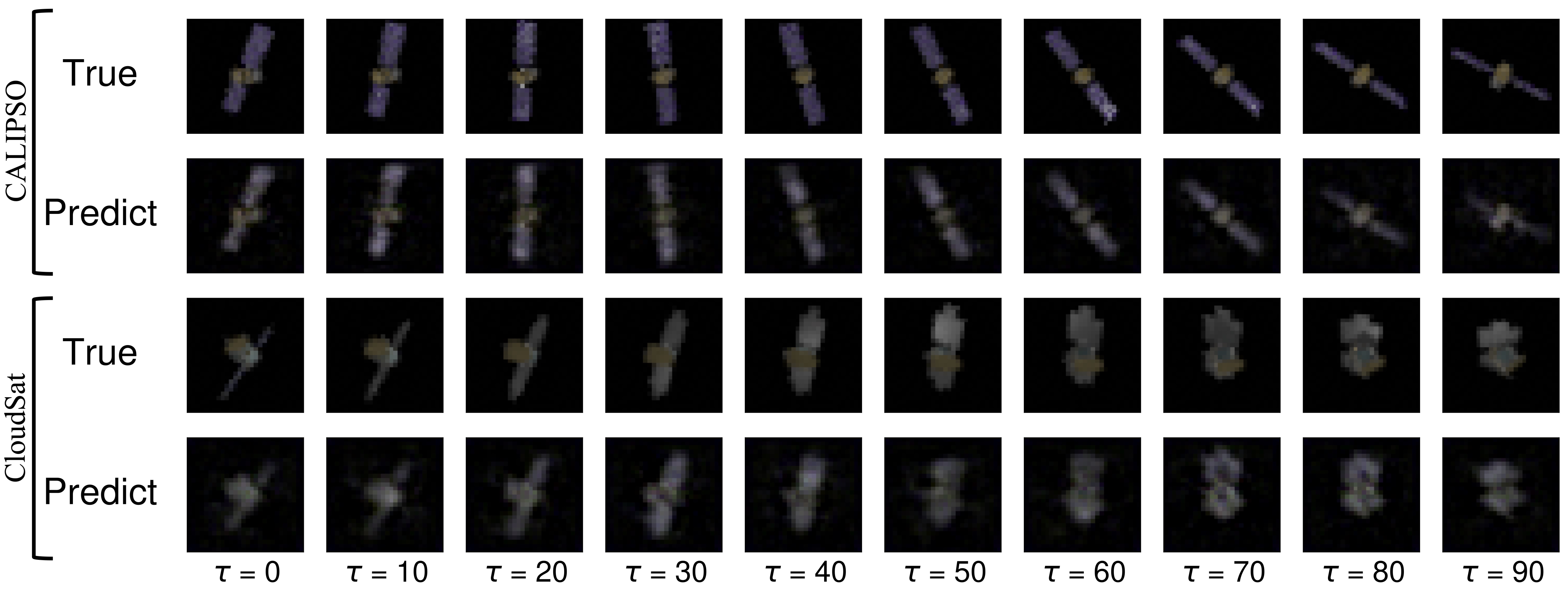}
  \caption{Predicted sequences for the CALIPSO satellite
 (top) and CloudSat satellite (bottom) with uniform mass densities given by our model. The~figure shows predicted images at every 10th time step from $\tau = $ 0 to 90.}
  \label{fig: sat_prediction}
\end{figure}

Figures~\ref{fig: prediction} and \ref{fig: sat_prediction} show the model's performance on the datasets for both short- and long-term predictions. Figure~\ref{fig: prediction} results show that the model is capable of predicting into the future at least five fold longer than the length of the time horizon used at training time. Figure~\ref{fig: sat_prediction} results show that the model is capable of predicting the future with images of more complex geometries and surface properties, i.e.,~those of the CALIPSO and CloudSat satellites, at~least ten fold longer than the length of the time horizon used at training time.
The model's performance on the datasets is indicative of its capabilities to predict dynamics and map them to image~space. 

\unskip

The uniform mass density cube and prism datasets are used to demonstrate baseline capabilities of our approach for image prediction. The~non-uniform mass density datasets validate the model's capability to predict a mass distribution that may not be visible from the exterior, e.g.,{\jjm{ for an asteroid or space debris or}}~as part of failure diagnostics on a satellite where there may be broken or shifted internal components. {\jjm{The satellite datasets are used to validate the model's capability to handle bodies with less regular and more realistic external geometries. }} 

We compare the performance of our model to three baseline models: (1) the {{Long Short Term Memory}} (LSTM) network, (2) the Neural ODE~\cite{chen2018neural} network, and~(3) the Hamiltonian Generative Network (HGN)~\cite{Toth2020HGN}. Recurrent neural networks like the LSTM-baseline provide a discrete dynamics model. Neural ODE can be combined with a multi-layer perceptron to predict continuous dynamics. HGN is a generative model with a Hamiltonian inductive bias. Architecture and training details for each baseline are given in Appendix~\ref{sec: baseline}. The~prediction performances of our model and the baselines are shown in Table~\ref{table: pixel-mse}. {{Our model has the lowest MSE on the majority of our datasets with good prediction performance on all of our datasets. Our model outperforms the state-of-the-art HGN model, reducing the expected MSE by nearly half on all datasets.}} Overall, our model outperforms the baseline models on the majority of the datasets with a more interpretable latent space, continuous dynamics, and~fewer model~parameters.

\begin{table}[H]
 \caption{Average pixel mean square error over a 30-step prediction on the train and test data on six datasets. All values are multiplied by 1e+3. We evaluate our model and compare to three baseline models: (1) recurrent model (LSTM~\cite{hochreiter1997long}), (2) Neural ODE (\cite{chen2018neural}), and (3) HGN (\cite{Toth2020HGN}).}
\resizebox{\textwidth}{!}{
  \begin{tabular}{lllllllll}
    \toprule
    \multirow{2}{*}{\textbf{Dataset}} &
      \multicolumn{2}{c}{\textbf{Ours}} &
      \multicolumn{2}{c}{\textbf{LSTM-Baseline}} &
      \multicolumn{2}{c}{\textbf{Neural ODE-Baseline}} &
      \multicolumn{2}{c}{\textbf{HGN}}\\
      & \multicolumn{1}{c}{\textbf{TRAIN}} & \multicolumn{1}{c}{\textbf{TEST}} & \multicolumn{1}{c}{\textbf{TRAIN}} & \multicolumn{1}{c}{\textbf{TEST}} & \multicolumn{1}{c}{\textbf{TRAIN}} & \multicolumn{1}{c}{\textbf{TEST}}
      & \multicolumn{1}{c}{\textbf{TRAIN}} & \multicolumn{1}{c}{\textbf{TEST}}\\
      \midrule
    Uniform Prism & \textbf{2.66 $\pm$ 0.10}  &  \textbf{2.71 $\pm$ 0.08} & 3.46 $\pm$ 0.59 & 3.47 $\pm$ 0.61 & 3.96 $\pm $ 0.68 & 4.00 $\pm$ 0.68 & 4.18 $\pm $ 0.0 & 7.80 $\pm$ 0.30\\
    Uniform Cube & \textbf{3.54 $\pm$ 0.17} &  \textbf{3.97 $\pm$ 0.16} & 21.55 $\pm$ 1.98 & 21.64 $\pm$  2.12 & 9.48 $\pm$ 1.19 & 9.43 $\pm$ 1.20 & 17.43 $\pm $ 0.00 & 18.69 $\pm$ 0.12\\
    Non-uniform Prism & \textbf{4.27 $\pm$ 0.18} & 6.61 $\pm$ 0.88 & 4.50 $\pm$ 1.31 & \textbf{4.52 $\pm$ 1.34} & 4.67 $\pm$ 0.58 & 4.75 $\pm$ 0.59 & 6.16 $\pm $ 0.08 & 8.33 $\pm$ 0.26\\
    Non-uniform Cube & \textbf{6.24 $\pm$ 0.29} & \textbf{4.85 $\pm$ 0.35} & 7.47 $\pm$ 0.51  & 7.51 $\pm$ 0.50 & 7.89 $\pm$ 1.50 & 7.94 $\pm$ 1.59 & 14.11 $\pm $ 0.13 & 18.14 $\pm$ 0.36\\
    CALIPSO & 0.79 $\pm$ 0.53 & 0.87 $\pm$ 0.50 & \textbf{0.62 $\pm$ 0.21}  & \textbf{0.65 $\pm$ 0.22} & 0.69 $\pm$ 0.26 & 0.71 $\pm$ 0.27 & 1.18 $\pm $ 0.02 & 1.34 $\pm$ 0.05\\
    CloudSat & \textbf{0.64 $\pm$ 0.45} & \textbf{0.65 $\pm$ 0.29} & 0.89 $\pm$ 0.36  & 0.93 $\pm$ 0.43 & 0.65 $\pm$ 0.22 & 0.66 $\pm$ 0.25 & 1.48 $\pm $ 0.04 & 1.66 $\pm$ 0.11\\
    \midrule
    Number of Parameters &  \multicolumn{2}{c}{6} &  \multicolumn{2}{c}{52,400} &  \multicolumn{2}{c}{11,400} & \multicolumn{2}{c}{-}\\ 
    \bottomrule
\end{tabular}%
 }
 \label{table: pixel-mse}
\end{table}

In Appendix~\ref{appendix: ablation}, we present the results of ablations studies and provide discussion. {\jjm{We find that the latent losses improve performance. However, the model may be over constrained with both the dynamics-based and auto-encoding based reconstruction losses.}}

\section{{Summary and~Conclusions}}
\label{sec: conclusion}
\unskip

\subsection{{Summary}}
In this work, we have presented the first physics-informed deep learning framework for predicting image sequences of 3D rotating rigid-bodies by embedding the images as measurements in the configuration space $\mathbf{SO}(3)$ and propagating the Hamiltonian dynamics forward in time. We have evaluated our approach on new datasets of free-rotating 3D bodies with different inertial properties, and~have demonstrated the ability to perform long-term image predictions. {We outperform the LSTM, Neural ODE and Hamiltonian Generative Network (HGN) baselines on our datasets, producing better qualitative predictions and reducing the error observed for the state-of-the-art HGN by a factor of 2.}

\subsection{{Conclusions}}
By enforcing the representation of the latent space to be $\mathbf{SO}(3)$, this work provides the advantage of interpretability over black-box physics-informed approaches. The~extra interpretability of our approach is a step towards placing additional trust into sophisticated deep learning models. This work provides a natural path to investigating how to incorporate and evaluate the effect of classical model-based control directly to trajectories in the latent space. 
\section{Future~Work}
Although our approach so far has been limited to embedding RGB-images of rotating rigid-bodies with configuration spaces in $\mathbf{SO}(3)$, there are natural extensions to a wider variety of problems. For~instance, this framework can be extended to embed different high-dimensional sensor measurements, such as point clouds, by~modifying the feature extraction layers of the autoencoder. The~latent space can be chosen to reflect generic rigid bodies in $\mathbf{SE}(3)$ or systems in more complicated spaces, such as the $n$-jointed robotic arm on a restricted subspace of $\Pi_{i=1}^{n}\left(\mathbf{SO}(3)\right)$. {{Another possible extension includes multibody systems, i.e.,~systems with rigid and flexible body dynamics, which would have applications to systems such as spacecraft with flexible solar panels and aircraft with flexible wings.}}

\vspace{6pt}

\authorcontributions{J.J.M. and C.A.-B. are the lead authors of this manuscript. Both assisted in conceptualization, methodology, investigation, writing, and editing of the manuscript. J.J.M. also contributed analyses and~led software development associated with the model and assisted in software development for data generation. C.A.-B. also contributed compute resources and assisted in software development, and~served in a supervisory role. N.Z. contributed to methodology, investigation, and~writing and editing of the manuscript, and~led data curation and software development for data generation. E.D. provided resources, served in a supervisory role, and~assisted in writing and editing of the manuscript. N.E.L. was the {\jjm{principal}} investigator, contributed to conceptualization, methodology, investigation, writing, and editing the manuscript, and~served in a supervisory role and acquired~funding. All authors have read and agreed to the published version of the manuscript.}

\funding{{\jjm{This research was funded in part by the Office of Naval Research grant number \#N00014-18-1-2873 and in part by funding provided by The Aerospace Corporation. The APC was funded by The Aerospace Corporation.}}}

\dataavailability{{\jjm{The code used to create and train our model is available \href{https://github.com/CAB-Lab-Princeton/Learning-RBD-from-Images}{here}. The dataset generation code is available \href{https://github.com/jjmason687/rbnn_data_generation}{here}.}}} 

\acknowledgments{Justice Mason and Christine Allen-Blanchette would like to thank Yaofeng Desmond Zhong and Juncal Arbelaiz for their helpful~discussions.}

\conflictsofinterest{The authors declare no conflicts of interest.}

\appendixtitles{no}
\appendixstart
\appendix
\section[\appendixname~\thesection]{}
\appendixtitles{yes}
\subsection[\appendixname~\thesubsection]{Rigid Body Rotational Dynamics and~Stability}
\label{app:rigid_body}
 Let $\mathbf{J} \in \mathbb{R}^{3 \times 3}$ denote the \textit{moment-of-inertia matrix} for a 3D rigid body. The~matrix $\mathbf{J}$ depends on how mass is distributed inside the body and can be understood to play a role in rotational dynamics that is analogous to, but~more complicated than, the~role played in translational dynamics of the scalar total body mass $m$.
 
Define an orthonormal reference frame ${\cal B} = \{\mathbf{i}, \mathbf{j}, \mathbf{k}\}$ fixed to the body with origin at the body's center of mass. Let $\mathbf{r} = (x, y, z)$ be a point on the body expressed with respect to ${\cal B}$. The~distribution of mass inside the rigid body is encoded by density $\rho(\mathbf{r})$, i.e.,~mass per unit volume of the body at the point $\mathbf{r}$. Let $V$ be the total volume of the body and denote by $\otimes$ the outer product. $\mathbf{J}$ is computed~\cite{GoldsteinCM} with respect to body frame ${\cal B}$ as
\begin{equation}
    \label{eq:MOI_eqn}
    \mathbf{J} = \iiint_{V} \rho(\mathbf{r}) (\lVert\mathbf{r}\rVert^{2} \mathbf{I}_{3} - \mathbf{r} \otimes \mathbf{r}) dxdydz.
\end{equation}

$\mathbf{J}$ is a symmetric positive definite matrix, which means that it can always be diagonalized. If~the frame ${\cal B}$ is chosen so that $\mathbf{J}$ is diagonal, the~axes of  ${\cal B}$ are called the \textit{principal axes} and the three diagonal elements of $\mathbf{J}$  are called the \textit{principal moments of inertia}.

Consider, for~example, the~rectangular prism of Figure~\ref{fig: mass-distr-illustration}a,b, which has uniformly distributed mass, i.e., $\rho(\mathbf{r}) =\rho_0$ for every point $\mathbf{r}$ in the body. Let ${\cal B}$ be chosen with its first, second, and~third axes aligned with the long, intermediate, and~short axes of the prism, respectively. Then, the axes of ${\cal B}$ are the {\jjm{principal}} axes, $\mathbf{J} = \mathbf{J}_1$ is diagonal, and~the first, second, and~third {\jjm{principal}} moments of inertia (the diagonal elements of $\mathbf{J}_1$) are ordered from smallest to largest. For~the very same rectangular prism but with the non-uniform distribution of mass used in Figure~\ref{fig: mass-distr-illustration}d, the~moment-of-inertia matrix $\mathbf{J}_3$, with~respect to the same frame ${\cal B}$, is no longer diagonal and its {\jjm{principal}} moments of inertia are different from those in the uniform case. $\mathbf{J}_1$ and $\mathbf{J}_3$ as well other moment-of-inertia matrices used for experiments in this work are given in Appendix~\ref{appendix: hyper-parameters}.

Let $\mathbf{\Omega}_{0} = \mathbf{\Omega}(0)$ be the initial body angular velocity. Euler's equations~\cite{GoldsteinCM} describe the rotational dynamics of the  body, i.e.,~the evolution over time $t$ of $\boldsymbol{\Pi}$ given $\mathbf{J}$ and $\mathbf{\Omega}_{0}$:
\begin{equation}
    \label{eq:EulerEqn}
    \frac{d \boldsymbol{\Pi}(t)}{dt} = \boldsymbol{\Pi}(t) \times \mathbf{J}^{-1}\boldsymbol{\Pi}(t), \;\;\; \mathbf{\Pi}(0) = \mathbf{J}\mathbf{\Omega}_{0},
\end{equation}
where $\times$ is the vector cross product. The~corresponding evolution of body angular velocity over time  is $\mathbf{\Omega}(t) = \mathbf{J}^{-1} \boldsymbol{\Pi}(t)$, where $\boldsymbol{\Pi}(t)$ is the solution of \eqref{eq:EulerEqn}. 

Given $\mathbf{\Omega}(t)$, $t \geq 0$, the~evolution of orientation over time is computed from the rigid body kinematics equations:
\begin{equation}
    \label{eq:Rdot}
\frac{d\mathbf{R(t)}}{dt} = \mathbf{R}(t)\; \boldsymbol{\Omega}_{\times}(t),
\end{equation}
where $\boldsymbol{\Omega}_{\times}$ is the $3\times3$ skew-symmetric matrix defined by $(\boldsymbol{\Omega}_{\times}) \boldsymbol{y} = \boldsymbol{\Omega} \times \boldsymbol{y}$ for $\boldsymbol{y} \in \mathbb{R}^3$. 

For the rotational dynamics \eqref{eq:EulerEqn}, there are three equilibrium solutions, i.e.,~where $d \boldsymbol{\Pi}(t)/dt = 0$, corresponding to steady spin about the short principal axis, intermediate principal axis, and~long principal axis, respectively. Steady spin about the short axis and long axis is stable, which means that an initial angular velocity near either of these solutions yields a spinning behavior, independent of exterior geometry (see Figure~\ref{fig: mass-distr-illustration}b,c). Steady spin about the intermediate axis is unstable, which means that an initial angular velocity near this solution yields a tumbling behavior (see Figure~\ref{fig: mass-distr-illustration}a).

Figure~\ref{fig: mass-distr-illustration}a,b shows that for the same prism with the same (uniform) mass distribution, and thus the same moment-of-inertia matrix $\mathbf{J}_1$, different values of initial body angular velocity result in very different behavior: an unstable tumble in Figure~\ref{fig: mass-distr-illustration}a and a stable spin in Figure~\ref{fig: mass-distr-illustration}b. Figure~\ref{fig: mass-distr-illustration}b,d shows that for the same prism with the same initial angular velocity, different mass distributions yield different behaviors, a~steady spin in (b) when $\mathbf{J} =\mathbf{J}_1$ and a wobble in (d) when $\mathbf{J} =\mathbf{J}_3$. Figure~\ref{fig: mass-distr-illustration}b,c shows that the rotational dynamics of a rigid body with the same moment-of-inertia matrix $\mathbf{J}_1$ and same initial body angular velocity yield the same behavior, despite different exterior geometries, i.e.,~the prism in (b) and the CALIPSO satellite in (d). 

These cases illustrate that without a way of inferring the underlying mass distribution and estimating initial conditions, there is no way to predict the dynamics from~images.
\subsection{Dataset Generation~Parameters}
\label{appendix: hyper-parameters}
\subsubsection{Uniform Mass Density Cube}
The moment-of-inertia tensor and its inverse for the uniform mass density cube are given by the matrices $\mathbf{J}_{0}$ and $\mathbf{J}_{0}^{-1}$ in Table~\ref{table: mass-props}. The~{\jjm{principal}} axes of rotation expressed in the body-fixed reference frame are also given in Table~\ref{table: mass-props}, showing the {\jjm{principal}} axes and body-fixed reference frame are~aligned.

\subsubsection{Uniform Mass Density Prism}
The moment-of-inertia tensor and its inverse for the uniform mass density prism are given by the matrices $\mathbf{J}_{1}$ and $\mathbf{J}_{1}^{-1}$ in Table~\ref{table: mass-props}. The~{\jjm{principal}} axes of rotation expressed in the body-fixed reference frame are also given in Table~\ref{table: mass-props}, showing the {\jjm{principal}} axes and body-fixed reference frame are~aligned.

\subsubsection{Non-Uniform Mass Density Cube}
The moment-of-inertia tensor and its inverse for the non-uniform mass density cube are given by the matrices $\mathbf{J}_{2}$ and $\mathbf{J}_{2}^{-1}$ in Table~\ref{table: mass-props}. The~{\jjm{principal}} axes of rotation expressed in the body-fixed reference frame are also given in Table~\ref{table: mass-props}, and are not aligned with body-fixed reference~frame.

\subsubsection{Non-Uniform Mass Density Prism}
The moment-of-inertia tensor and its inverse for the non-uniform mass density prism are given by the matrices $\mathbf{J}_{3}$ and $\mathbf{J}_{3}^{-1}$ in Table~\ref{table: mass-props}. The~{\jjm{principal}} axes of rotation expressed in the body-fixed reference frame are also given in Table~\ref{table: mass-props}, and are not aligned with body-fixed reference~frame.

\subsubsection{CALIPSO}
The moment-of-inertia tensor and its inverse for the CALIPSO satellite are given by the matrices $\mathbf{J}_{4}$ and $\mathbf{J}_{4}^{-1}$ in Table~\ref{table: mass-props}. The~{\jjm{principal}} axes of rotation expressed in the body-fixed reference frame are also given in Table~\ref{table: mass-props}, i.e.,~the {\jjm{principal}} axes and body-fixed reference frame are~aligned.

\subsubsection{CloudSat}
The moment-of-inertia tensor and its inverse for the CloudSat satellite are given by the matrices $\mathbf{J}_{5}$ and $\mathbf{J}_{5}^{-1}$ in Table~\ref{table: mass-props}. The~{\jjm{principal}} axes of rotation expressed in the body-fixed reference frame are also given in Table~\ref{table: mass-props}, i.e.,~the {\jjm{principal}} axes and body-fixed reference frame are~aligned.

\begin{table}[H]
  \caption{Table containing the moment-of-inertia tensors, inverse moment-of-inertia tensors, and~{\jjm{principal}} axes used to generate training data for each~object.}
  \tablesize{\scriptsize}
  \label{table: mass-props}
\setlength{\cellWidtha}{\textwidth/4-2\tabcolsep-0.35in}
\setlength{\cellWidthb}{\textwidth/4-2\tabcolsep+0.05in}
\setlength{\cellWidthc}{\textwidth/4-2\tabcolsep+0.3in}
\setlength{\cellWidthd}{\textwidth/4-2\tabcolsep+0in}
\scalebox{1}[1]{\begin{tabularx}{\textwidth}{>{\raggedright\arraybackslash}m{\cellWidtha}>{\centering\arraybackslash}m{\cellWidthb}>{\centering\arraybackslash}m{\cellWidthc}>{\centering\arraybackslash}m{\cellWidthd}}
    \toprule
  {\textbf{Object}} & \textbf{Moment-of-Inertia Tensor} & \textbf{Inverse Moment-of-Inertia Tensor} &\textbf{{\jjm{principal}} Axes}\\
    \midrule
    Uniform Cube & $\mathbf{J}_{0} = \frac{1}{3}\Big(\begin{smallmatrix}1 & 0 & 0 \\
                                        0 & 1 & 0 \\
                                        0 & 0 & 1\end{smallmatrix}\Big)$ & $\mathbf{J}_{0}^{-1} = \Big( \begin{smallmatrix}3 & 0 & 0 \\
                                0 & 3 & 0 \\
                                0 & 0 & 3\end{smallmatrix} \Big)$ & $\Big\{\Big(\begin{smallmatrix}
                                            1\\ 0\\ 0
                                        \end{smallmatrix}\Big), \Big(\begin{smallmatrix}
                                            0\\ 1\\ 0
                                        \end{smallmatrix}\Big), \Big(\begin{smallmatrix}
                                            0\\ 0\\ 1
                                        \end{smallmatrix}\Big)\Big\}$ \\
                                        \\
    Uniform Prism & $\mathbf{J}_{1} = \Big(\begin{smallmatrix}0.42 & 0 & 0. \\
                                        0 & 1.41 & 0 \\
                                        0 & 0 & 1.67\end{smallmatrix}\Big)$ & $\mathbf{J}_{1}^{-1} = \Big(\begin{smallmatrix}2.40 & 0 & 0 \\
                                        0 & 0.71 & 0 \\
                                        0 & 0 & 0.60\end{smallmatrix} \Big)$ & $\Big\{\Big(\begin{smallmatrix}
                                            1\\ 0\\ 0
                                        \end{smallmatrix}\Big), \Big(\begin{smallmatrix}
                                            0\\ 1\\ 0
                                        \end{smallmatrix}\Big), \Big(\begin{smallmatrix}
                                            0\\ 0\\ 1
                                        \end{smallmatrix}\Big)\Big\}$ \\
                                        \\
    Non-uniform Cube & $\mathbf{J}_{2} = \Big(\begin{smallmatrix}0.17 & 0 & -0.56 \\
                                    0 & 0.17 & -0.99 \\
                                    -0.56 & -0.99 & 0.17\end{smallmatrix}\Big)$ & $\mathbf{J}_{2}^{-1} = \Big(\begin{smallmatrix}4.53 & -2.62 & -0.44 \\
                                    -2.62 &  1.34 & -0.78 \\
                                    -0.44 & -0.78 &-0.13\end{smallmatrix}\Big)$ & $\Big\{\Big(\begin{smallmatrix}
                                    -0.35\\ -0.62\\ -0.71
                                    \end{smallmatrix}\Big), \Big(\begin{smallmatrix}
                                    -0.87\\ 0.49\\ 0
                                    \end{smallmatrix}\Big), \Big(\begin{smallmatrix}
                                    -0.35\\ -0.62\\ 0.71
                                    \end{smallmatrix}\Big)\Big\}$ \\
                                    \\
    Non-uniform Prism & $\mathbf{J}_{3} = \Big(\begin{smallmatrix}0.47 & 0 & -0.28 \\
                                    0 & 1.61 & -0.49 \\
                                    -0.28 & -0.49 & 1.83\end{smallmatrix}\Big)$ & $\mathbf{J}_{3}^{-1} = \Big(\begin{smallmatrix}
                                    2.37 & 0.12 & 0.39 \\
                                    0.12 & 0.68 & 0.20 \\
                                    0.39 & 0.20 & 0.66\end{smallmatrix}\Big)$ & $\Big\{\Big(\begin{smallmatrix}
                                    -0.35\\ -0.62\\ -0.71
                                    \end{smallmatrix}\Big), \Big(\begin{smallmatrix}
                                    -0.87\\ 0.49\\ 0
                                    \end{smallmatrix}\Big), \Big(\begin{smallmatrix}
                                    -0.35\\ -0.62\\ 0.71
                                    \end{smallmatrix}\Big)\Big\}$ \\
                                    \\
    CALIPSO & $\mathbf{J}_{4} = \Big(\begin{smallmatrix}0.33 & 0 & 0 \\
                                        0 & 0.50 & 0 \\
                                        0 & 0 & 1.0\end{smallmatrix}\Big)$ & $\mathbf{J}_{4}^{-1} = \Big(\begin{smallmatrix}3.0 & 0 & 0 \\
                                        0 & 2.0 & 0 \\
                                        0 & 0 & 1.0\end{smallmatrix} \Big)$ & $\Big\{\Big(\begin{smallmatrix}
                                            1\\ 0\\ 0
                                        \end{smallmatrix}\Big), \Big(\begin{smallmatrix}
                                            0\\ 1\\ 0
                                        \end{smallmatrix}\Big), \Big(\begin{smallmatrix}
                                            0\\ 0\\ 1
                                        \end{smallmatrix}\Big)\Big\}$ \\
                                        \\
    CloudSat & $\mathbf{J}_{5} = \Big(\begin{smallmatrix}0.33 & 0 & 0 \\
                                        0 & 0.50 & 0 \\
                                        0 & 0 & 1.0\end{smallmatrix}\Big)$ & $\mathbf{J}_{5}^{-1} = \Big(\begin{smallmatrix}3.0 & 0 & 0 \\
                                        0 & 2.0 & 0 \\
                                        0 & 0 & 1.0\end{smallmatrix} \Big)$ & $\Big\{\Big(\begin{smallmatrix}
                                            1\\ 0\\ 0
                                        \end{smallmatrix}\Big), \Big(\begin{smallmatrix}
                                            0\\ 1\\ 0
                                        \end{smallmatrix}\Big), \Big(\begin{smallmatrix}
                                            0\\ 0\\ 1
                                        \end{smallmatrix}\Big)\Big\}$ \\
                                        \\
    \bottomrule
  \end{tabularx}
}
\end{table}
\unskip

\subsection{Hyperparameters}
\label{appendix: training_params}
The hyperparameters used to train our model are given in Table~\ref{table: train_params}.{\jjm{Hyperparameters, distinct from model parameters, control the training process. We optimize over the model parameters using the Adam optimizer \cite{kingma2014adam}}}.

\begin{table}[H]
  \caption{Hyperparameters used to train model for the non-uniform mass density prism experiment. Only values differing from default values are given in~the table.}
  \label{table: train_params}
    \tablesize{\small}
 \setlength{\cellWidtha}{\textwidth/2-2\tabcolsep-0.5in}
\setlength{\cellWidthb}{\textwidth/2-2\tabcolsep+0.5in}
\scalebox{1}[1]{\begin{tabularx}{\textwidth}{>{\raggedright\arraybackslash}m{\cellWidtha}>{\raggedright\arraybackslash}m{\cellWidthb}}
    \toprule
    \multicolumn{2}{l}{\textbf{Experiment Hyperparameters}} \\
    \midrule
    \textbf{Parameter Name} & \textbf{Value} \\
    \midrule
    Random seed & 0 \\
    Test dataset split & 0.2 \\
    Validation dataset split & 0.1 \\
    Number of epochs & 1000 \\
    Batch size & 256 \\
    Autoencoder learning rate & $1\times10^{-3}$\\
    Dynamics learning rate & $1\times10^{-3}$\\
    Sequence length & 10\\
    Time step & $1\times10^{-3}$\\
    \bottomrule
  \end{tabularx}}
\end{table}

\subsection{Performance of Baseline~Models}\label{sec: baseline}
We compare the performance of our model against three baseline architectures: (1) LSTM, (2) Neural ODE~\cite{chen2018neural}, and~(3) Hamiltonian Generative Network (HGN)~\cite{Toth2020HGN}. LSTM and Neural ODE baselines are trained using the same autoencoder architecture as our model, 
while HGN is trained with the autoencoder architecture described in~\citet{Toth2020HGN}. The {\jjm{LSTM- and Neural ODE-baseline}} differ from our approach in how the dynamics are computed, emphasizing the beneficial role of Hamiltonian structure as well as our $\mathbf{SO}(3)$ latent space. 
\subsubsection{LSTM-Baseline}
\begin{figure}[H]
  \includegraphics[width=0.995\textwidth]{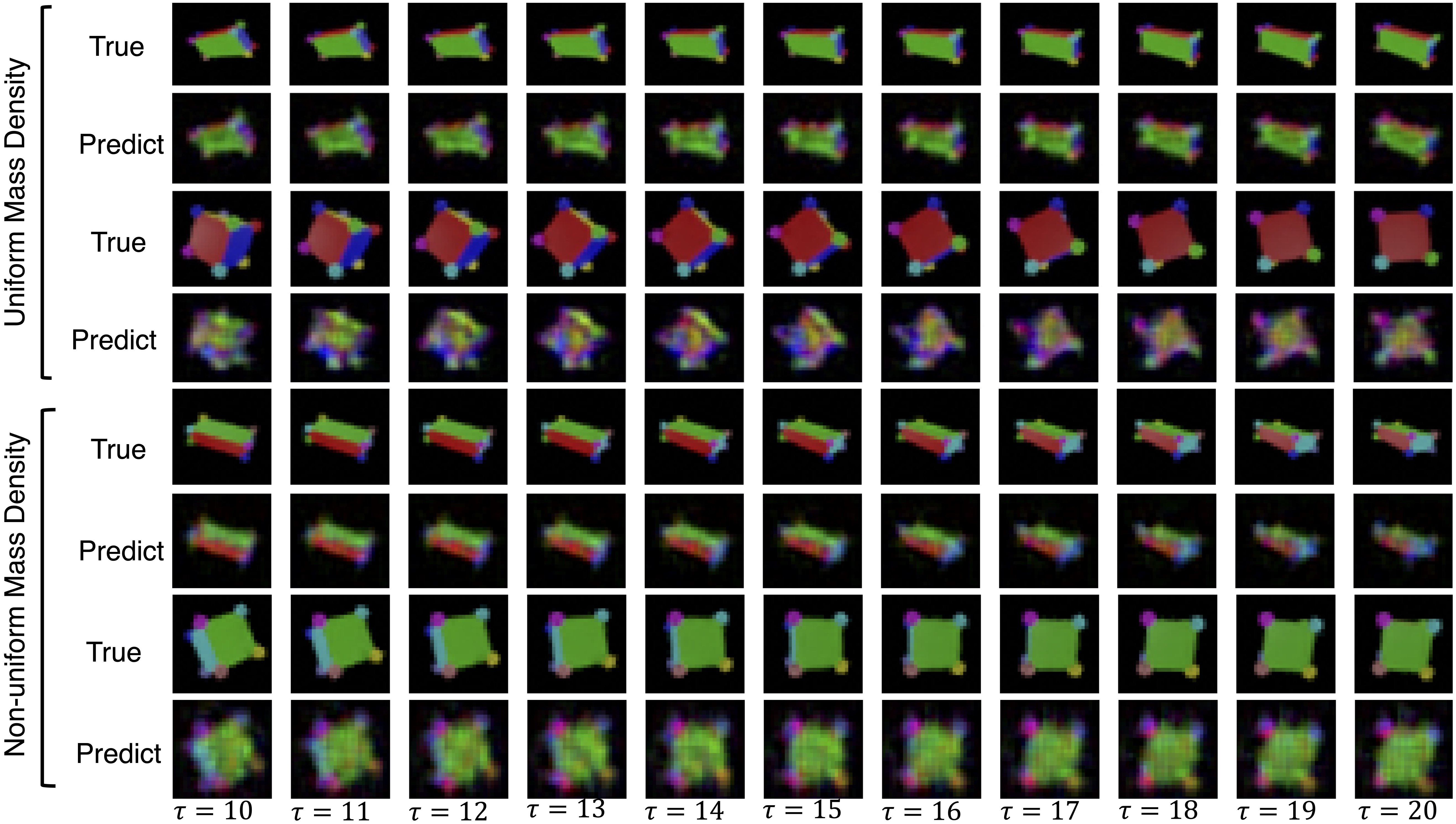}
  \caption{Predicted sequences for uniform/non-uniform prism and cube datasets given by the LSTM-baseline. 
  The figure shows time steps $\tau = 10$ through $\tau = 20$. These are the first 11 predictions of the model. }
  \label{fig: lstm-prediction}
\end{figure}

The LSTM-baseline uses an LSTM network to predict dynamics {\jjm{in the latent space}}. The~LSTM-baseline is a three-layer LSTM network with an input dimension of 6 and a hidden dimension of 50. The~hidden state and cell state are randomly initialized and the output of the network is mapped to a six-dimensional latent vector by a linear layer. The~LSTM-baseline predicts a single step forward using the nine previous states as input. We train the LSTM by minimizing the sum of the {\jjm{auto-encoding}} and dynamics{\jjm{-based reconstruction}} losses, $\mathcal{L}_\text{ae}$ and $\mathcal{L}_\text{dyn}$ as defined in Section~\ref{subsec: train_method}. 
At inference, we use a recursive strategy to predict farther into the future using previously predicted states to predict {\jjm{subsequent states}}. The~qualitative performance of the LSTM-baseline is given in Figure~\ref{fig: lstm-prediction}, and~the quantitative performance is given in Table~\ref{table: pixel-mse}. The~total number of parameters in the network is 52,400. The~LSTM-baseline has poorer performance than our proposed approach on all evaluated~datasets.

\subsubsection{Neural ODE~\cite{chen2018neural}-Baseline}
\begin{figure}[H]
  \includegraphics[width=0.995\textwidth]{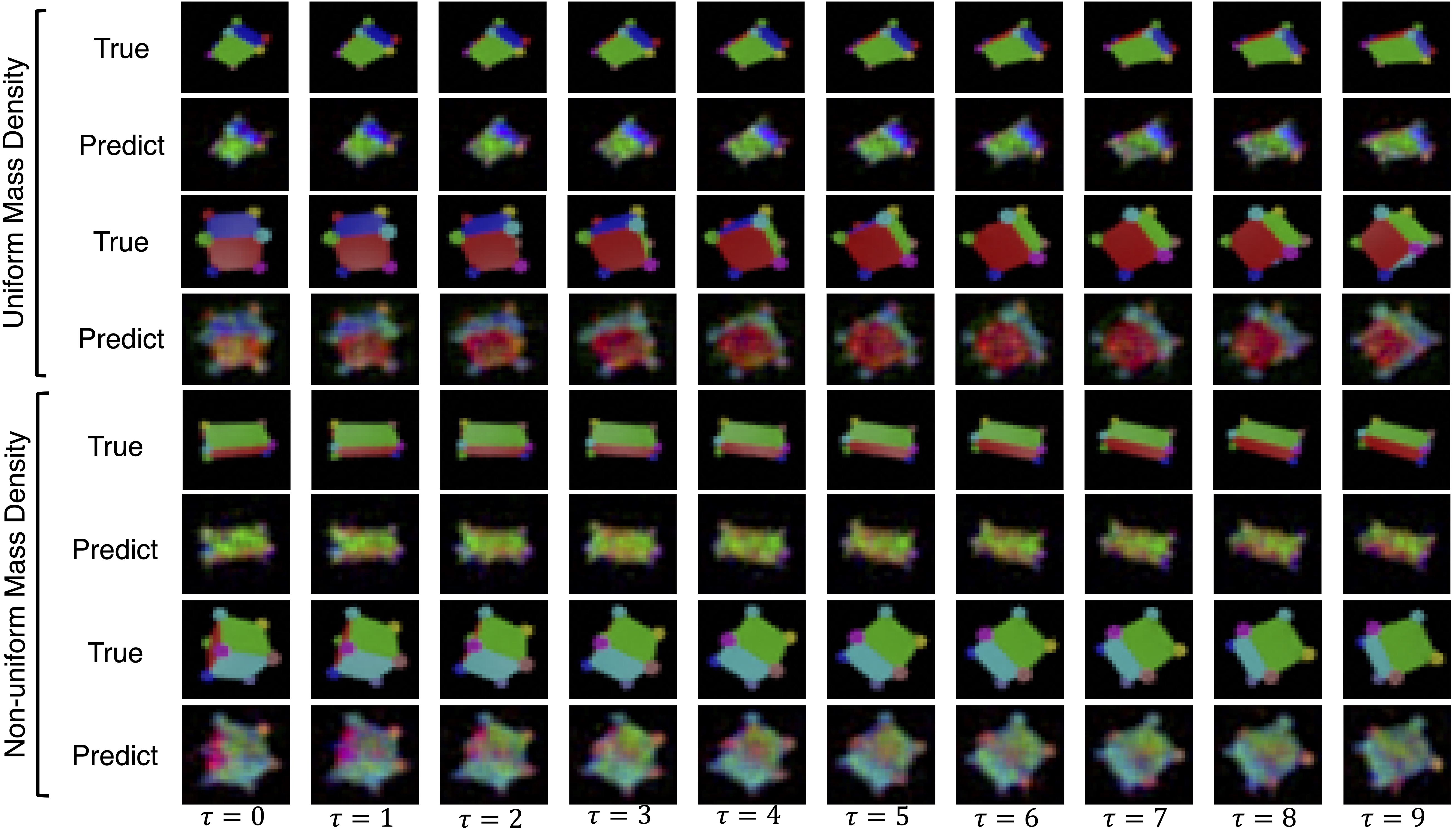}
  \caption{Predicted sequences for uniform/non-uniform prism and cube datasets given by the Neural ODE-baseline. 
  }
  \label{fig: node-prediction}
\end{figure}

The Neural ODE-baseline uses the Neural ODE~\cite{chen2018neural} framework {\jjm{to predict dynamics in the latent space}}. The~Neural ODE-baseline is a three-layer multilayer perceptron (MLP) that uses the ELU~\cite{clevert2015fast} nonlinear activation function. The~baseline has an input dimension of 6, a~hidden dimension of 50, and~an output dimension of 6. The~Neural ODE-baseline predicts a sequence of latent states using a single initial latent state. We train the Neural ODE-baseline by minimizing the sum of {\jjm{the auto-encoding and dynamics-based reconstruction}} losses, $\mathcal{L}_\text{ae}$ and $\mathcal{L}_\text{dyn}$, as defined in Section~\ref{subsec: train_method}. We use the RK4-integrator to integrate the learned dynamics. The~qualitative performance for the Neural ODE-baseline is given in Figure~\ref{fig: node-prediction}, and~the quantitative performance is given in Table~\ref{table: pixel-mse}. The~total number of parameters in the network is 11,406. The~Neural ODE-baseline has poorer performance than our proposed approach on all evaluated~datasets.

\unskip
\subsubsection{Hamiltonian Generative Network (HGN)}
\begin{figure}[H]
  \includegraphics[width=0.8\textwidth]{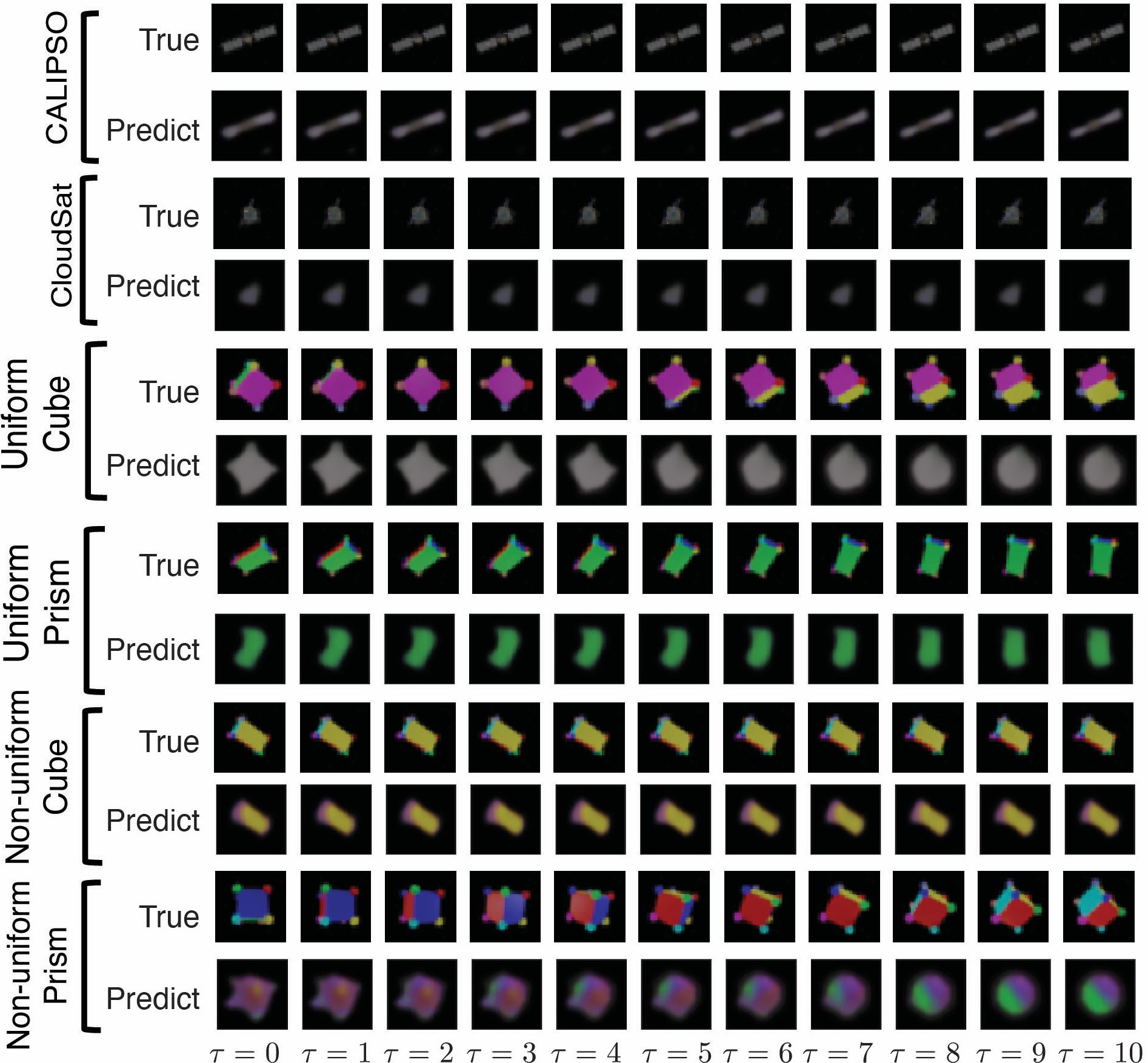}
  \caption{{\jjm{Predicted sequences for all datasets given by the Hamiltonian Generative Network (HGN) baseline. The figure shows timesteps $\tau = 0$ through $\tau=10$.}}}
  \label{fig: hgn-prediction}
\end{figure}

 HGN~\cite{Toth2020HGN} uses a combination of a variational auto-encoding (VAE) neural network and~Hamiltonian dynamics to perform video prediction. When testing HGN as a baseline, we use the implementation provided by \citet{balsells_rodas_carles_2021_4835278}. We train HGN on our datasets using {\jjm{the hyperparameters}}, loss function, and integrator {\jjm{described in~\citet{Toth2020HGN}. The~qualitative performance for the HGN-baseline is given in Figure~\ref{fig: hgn-prediction}}}, and the~quantitative performance is given in Table~\ref{table: pixel-mse}. 

\subsection{Ablation~Studies}
\label{appendix: ablation} 
\begin{figure}[H]
  \includegraphics[width=0.88\textwidth]{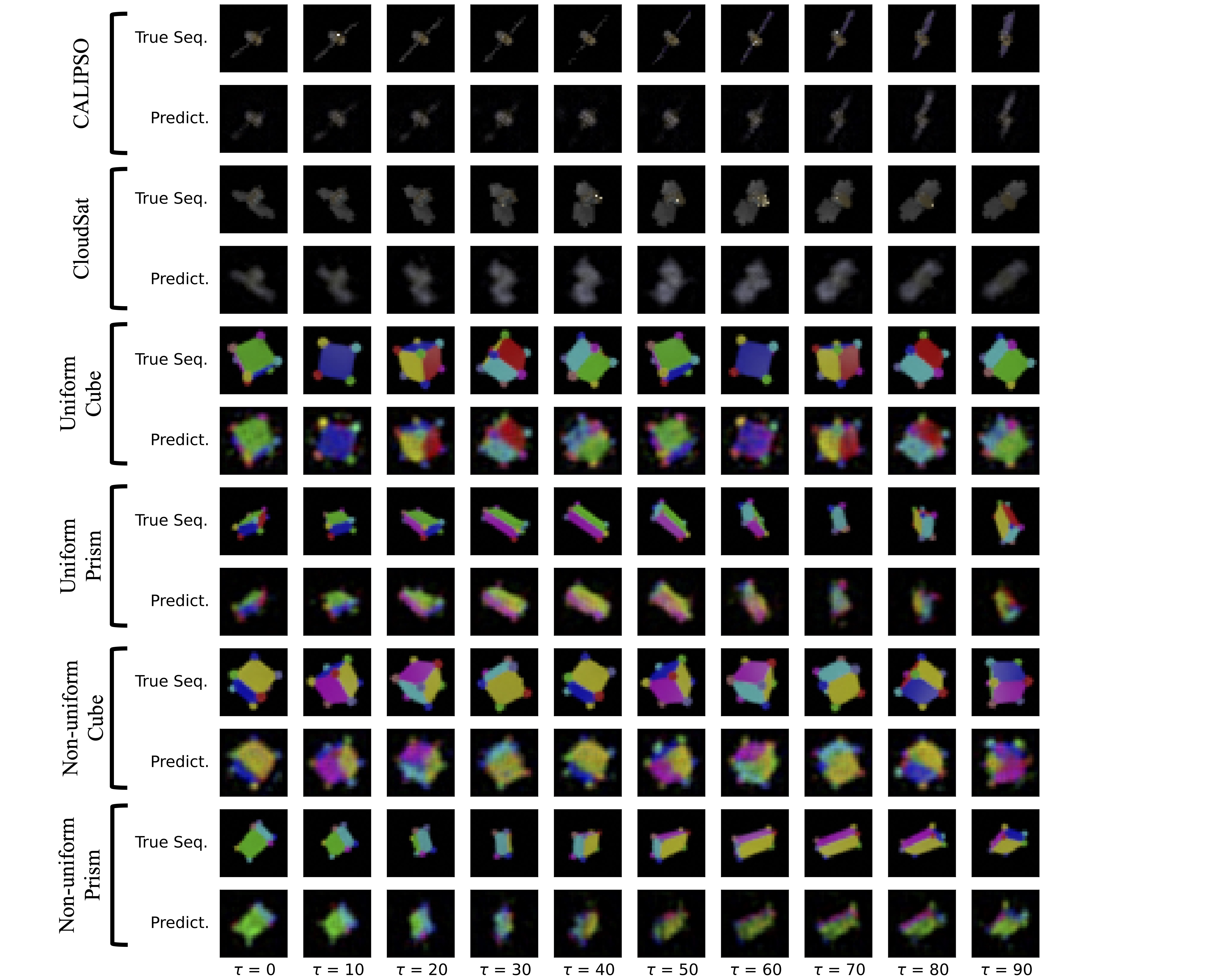}
  \caption{Evaluation of the image prediction performance of an ablated version of our model trained with only the {\jjm{dynamics-based reconstruction}} loss ($\mathcal{L}_\text{dyn}$) from Section~\ref{subsec:recon_losses}. The ablated model has poorer performance than the proposed approach over all datasets. The~prediction performance worsens earlier than the proposed model's performance, as shown in Figure~\ref{fig: prediction}.}
  \label{fig: ablation-pred-prediction}
\end{figure}

In our ablation studies, we explore the impact of the reconstruction losses and latent losses on the performance of our model (see Section~\ref{subsec:recon_losses} and ~\ref{subsec:latent_losses} for the definition of our losses). The~ablated models are trained similarly to the proposed model, {but parts of the loss functions are removed}. In~the first ablation study, only the dynamics-based reconstruction loss {\jjm{($\mathcal{L}_\text{dyn}$)}} is used {\jjm{, i.e., the auto-encoding reconstruction loss ($\mathcal{L}_\text{ae}$) and~latent losses (${\mathcal{L}_\text{latent}}_R$ and ${\mathcal{L}_\text{latent}}_\Pi$) are removed from the total loss function.}} In~the second ablation study,{\jjm{ only the auto-encoding reconstruction and dynamics-based reconstruction losses ($\mathcal{L}_\text{ae}$ and $\mathcal{L}_\text{dyn}$) 
 are used, i.e., }}the~latent losses {\jjm{(${\mathcal{L}_\text{latent}}_R$ and ${\mathcal{L}_\text{latent}}_\Pi$)}} are removed from the total loss function. {\jjm{In the final ablation study, only the dynamics-based reconstruction and latent losses are used ($\mathcal{L}_\text{dyn}$, ${\mathcal{L}_\text{latent}}_R$, and ${\mathcal{L}_\text{latent}}_\Pi$), i.e., the auto-encoding reconstruction loss ($\mathcal{L}_\text{ae}$) is removed from the total loss function.}} These ablation studies {\jjm{demonstrate the prediction performance of the model when trained with (1) the dynamics-based reconstruction}} loss only, (2) the auto-encoding reconstruction and dynamics-based reconstruction losses{\jjm{, and (3) the dynamics-based reconstruction loss and latent losses}}. In the first two cases of the ablation study, the~prediction performance worsens{\jjm{ on the majority of the datasets, but in the third case, the~prediction performance improves over the proposed model on the majority of the datasets. It may be that model is over constrained with both the dynamics-based and auto-encoding based reconstruction losses.}}
 From Table~\ref{table: pixel-mse-ablation} and Figure~\ref{fig: ablation-pred-prediction}, it can be inferred that only using the {\jjm{dynamics-based reconstruction}} loss negatively affects the prediction performance of our proposed model (although it is still 
 better than the baselines in Table~\ref{table: pixel-mse}). 

\begin{figure}[H]
  \includegraphics[width=0.8\textwidth]{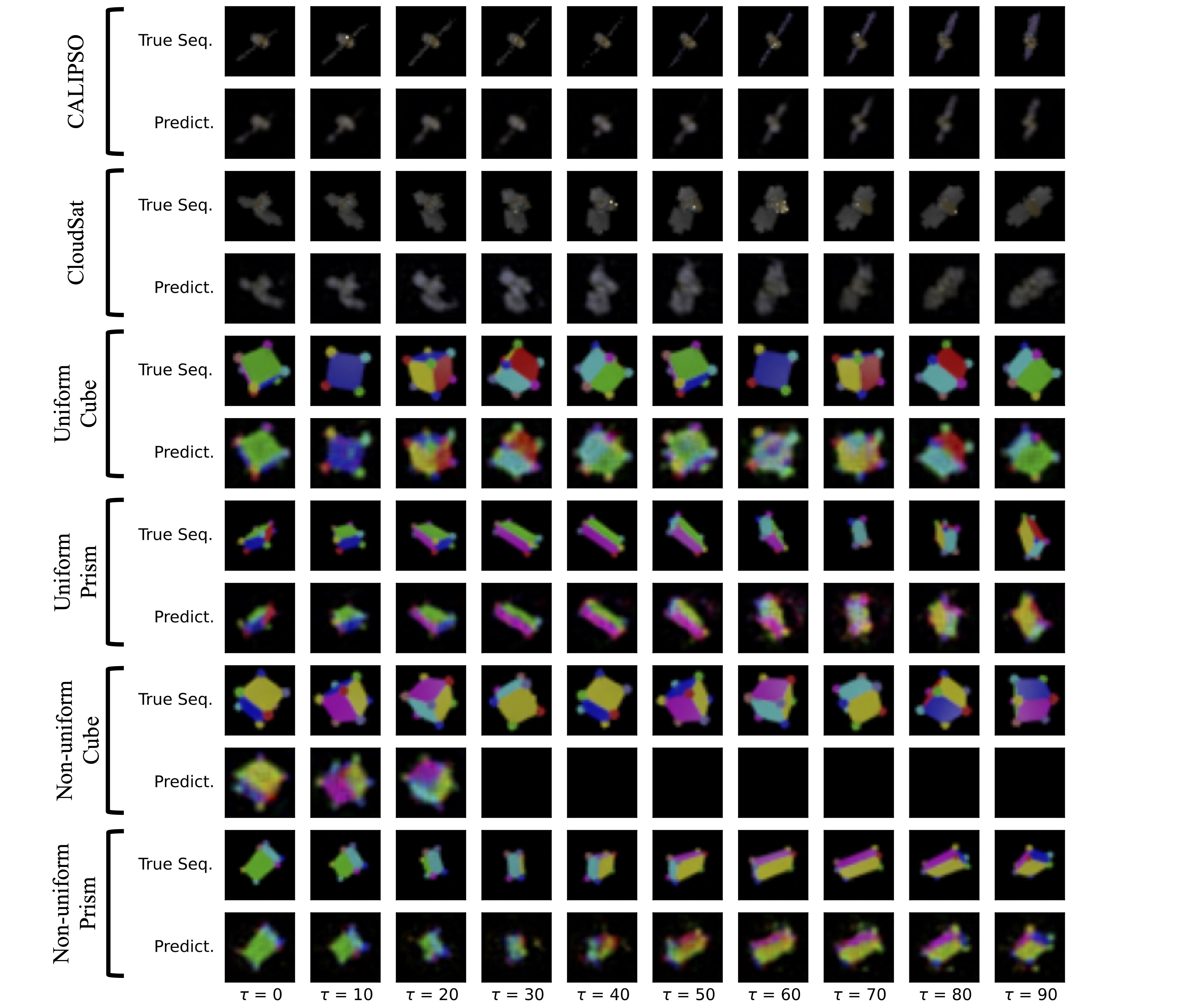}
  \caption{Evaluation of the image prediction performance of an ablated version of our model trained {\jjm{with only the auto-encoding reconstruction and dynamics-based reconstruction losses ($\mathcal{L}_\text{ae}$ and $\mathcal{L}_\text{dyn}$) from Section~\ref{subsec:recon_losses}.}} The ablated model has poorer performance than the proposed approach over all datasets---even failing to predict after $\sim$30 time steps for the non-uniform cube dataset.
  }
  \label{fig: ablation-latent-prediction}
\end{figure}

 Table~\ref{table: pixel-mse-ablation} and Figure~\ref{fig: ablation-latent-prediction} {\jjm{also demonstrate the positive impact}} of the latent losses on prediction performance for our model. We see worsened average pixel MSE and prediction{\jjm{ with the latent losses removed}}. These results are further corroborated in the literature~\cite{Allen-Blanchette2020Lag, WatterManuelE2C}. {\jjm{Furthermore, Figure~\ref{fig: ablation-ae-prediction} show a better prediction performance when the auto-encoding reconstruction loss is removed. This could indicate that the $\mathcal{L}_\text{ae}$ loss is over-constraining the proposed model.}}
\unskip
\unskip
{\jjm{\begin{figure}[H]
  \includegraphics[width=0.8\textwidth]{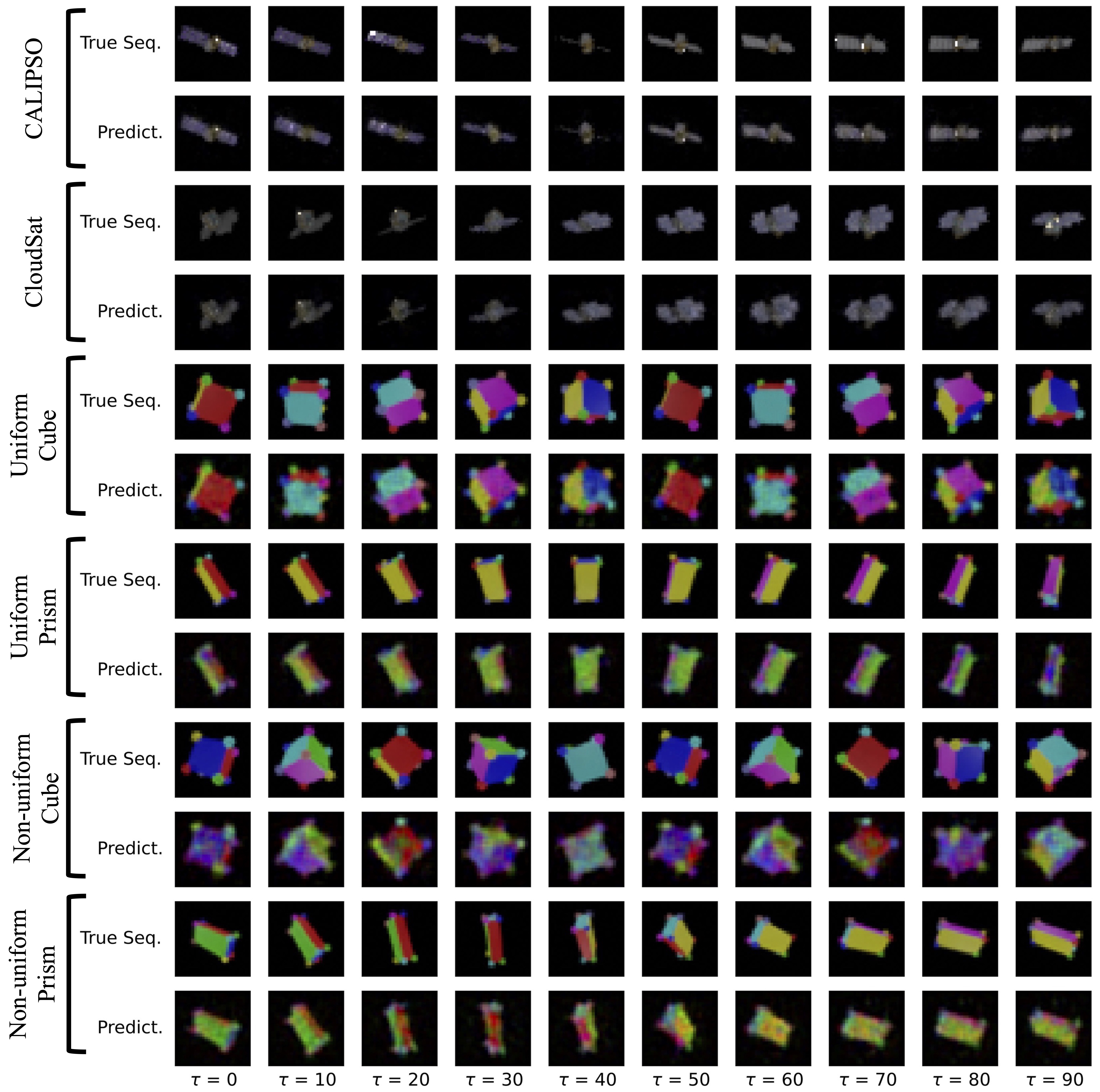}
  \caption{{\jjm{Evaluation of the image prediction performance of an ablated version of our model trained with only dynamics-based reconstruction losses and latent losses ($\mathcal{L}_\text{dyn}$, ${\mathcal{L}_\text{latent}}_R$, and ${\mathcal{L}_\text{latent}}_\Pi$) from Section~\ref{subsec:recon_losses}. The ablated model has better performance than the proposed approach on the majority of the datasets-- implying that the proposed model may be over constrained.}}
  }
  \label{fig: ablation-ae-prediction}
\end{figure}
\unskip}}
\begin{table}[H]
 \caption{Average pixel MSE over a 30-step unroll on the train and test data on four datasets for our ablation study.
 }
\resizebox{\columnwidth}{!}{
  \begin{tabular}{lllllllllll}
    \toprule
    \multirow{2}{*}{\textbf{Dataset}} &
      \multicolumn{2}{c}{\boldmath{$\mathcal{L}$}\textsubscript{\textbf{total}}} &
      \multicolumn{2}{c}{\boldmath{$\mathcal{L}$}\textsubscript{\textbf{dyn}}} &
      \multicolumn{2}{c}{\boldmath{$\mathcal{L}$}\textsubscript{\textbf{dyn}} + \boldmath{$\mathcal{L}$}\textsubscript{\textbf{ae}}} &
      \multicolumn{2}{c}{{\jjm{\boldmath{$\mathcal{L}$}\textsubscript{\textbf{dyn}} + \boldmath{$\mathcal{L}$}\textsubscript{\textbf{latent}}}}}\\
      & \multicolumn{1}{c}{\textbf{TRAIN}} & \multicolumn{1}{c}{\textbf{TEST}} & \multicolumn{1}{c}{\textbf{TRAIN}} & \multicolumn{1}{c}{\textbf{TEST}} & \multicolumn{1}{c}{\textbf{TRAIN}} & \multicolumn{1}{c}{\textbf{TEST}} & \multicolumn{1}{c}{{\jjm{\textbf{TRAIN}}}} & \multicolumn{1}{c}{{\jjm{\textbf{TEST}}}} \\
      \midrule
    Uniform Prism & \textbf{3.03 $\pm$ 1.26} & \textbf{3.05 $\pm$ 1.21} & 3.99 $\pm$ 1.21 & 3.74 $\pm$ 0.93 & 3.99 $\pm$ 1.50 & 3.85 $\pm$ 1.45 & {\jjm{4.82 $\pm$ 1.32}} & {\jjm{5.09 $\pm$ 1.53}}\\
    Uniform Cube & 4.13 $\pm$ 2.14 & 4.62 $\pm$ 2.02 & 5.73 $\pm$ 0.51 & 5.87 $\pm$ 0.56 & 7.11 $\pm$ 2.63 & 6.95 $\pm$ 2.41 & {\jjm{\textbf{2.80 $\pm$ 0.18}}} & {\jjm{\textbf{2.80 $\pm$ 0.20}}}\\
    Non-uniform Prism & 4.98 $\pm$ 1.26 & 7.07 $\pm$ 1.88 & 4.27 $\pm$ 1.28 & 3.89 $\pm$ 1.10 & {{\textbf{3.86 $\pm$ 1.38}}} & {{\textbf{3.66 $\pm$ 1.27}}} & {\jjm{4.16 $\pm$ 1.27}} & {\jjm{5.09 $\pm$ 1.53}}\\
    Non-uniform Cube & 7.27 $\pm$ 1.06 & \textbf{5.65 $\pm$ 1.50} & \textbf{6.23 $\pm$ 0.88} & 5.93 $\pm$ 0.85 & -  & - & {\jjm{8.78 $\pm$ 0.93}} & {\jjm{8.64 $\pm$ 1.14}} \\
    CALIPSO & 1.18 $\pm$ 0.43 & 1.19 $\pm$ 0.63 & 2.00 $\pm$ 0.78  & 1.85 $\pm$ 0.58 & 1.73 $\pm$ 0.73 & 1.62 $\pm$ 0.50 & {\jjm{\textbf{0.49 $\pm$ 0.07}}} & {\jjm{\textbf{0.54 $\pm$ 0.18}}}\\
    CloudSat & 1.32 $\pm$ 0.74 & 1.56 $\pm$ 1.01 & 0.96 $\pm$ 0.17  & 1.39 $\pm$ 0.48 & 0.87 $\pm$ 0.29 & 1.40 $\pm$ 0.40 & {\jjm{\textbf{0.28 $\pm$ 0.06}}} & {\jjm{\textbf{0.28 $\pm$ 0.06}}}\\
    \bottomrule
  \end{tabular}%
 }
 \label{table: pixel-mse-ablation}
\end{table}

\begin{adjustwidth}{-\extralength}{0cm}

\reftitle{References}

\PublishersNote{}
\end{adjustwidth}
\end{document}